\documentclass{article} 
\usepackage{iclr2027_conference,times}

\usepackage{amsmath,amsfonts,bm}

\def\eqref#1{equation~\ref{#1}}

\def\1{\bm{1}}

\DeclareMathAlphabet{\mathsfit}{\encodingdefault}{\sfdefault}{m}{sl}
\SetMathAlphabet{\mathsfit}{bold}{\encodingdefault}{\sfdefault}{bx}{n}

\DeclareMathOperator*{\argmin}{arg\,min}

\usepackage{hyperref}
\usepackage{url}

\usepackage{tikz}
\usetikzlibrary{arrows.meta,calc}
\usepackage{figures/palette}
\usepackage{wrapfig}
\usepackage{enumitem}
\usepackage{amsmath}
\usepackage{amssymb}
\usepackage{booktabs}
\usepackage{array}
\usepackage{multirow}
\usepackage{xspace}
\usepackage[ruled,vlined,linesnumbered]{algorithm2e}
\SetAlgoNlRelativeSize{0}                 
\makeatletter  
\renewcommand{\algocf@makecaption@ruled}[2]{%
\global\sbox\algocf@capbox{\hskip\AlCapHSkip%
    \setlength{\algocf@lcaptionbox}{\algowidth}\addtolength{\algocf@lcaptionbox}{-2\AlCapHSkip}%
    \parbox[t]{\algocf@lcaptionbox}{\algocf@captiontext{#1}{#2}}}}%
\makeatother
\usepackage{tabularx}
\usepackage{arydshln}
\usepackage{csquotes}
\usepackage{tabularx}
\usepackage{arydshln}
\usepackage{wrapfig}
\usepackage{float}
\usepackage{xspace}
\usepackage{caption}
\usepackage{graphicx}
\usepackage{amsthm}
\usepackage[table]{xcolor}
\usepackage{listings}

\newcommand{\lever}{\textsc{Lever}\xspace}
\newcommand{\ie}{\textit{i.e.}}

\newcommand{\eg}{\textit{e.g.}}
\newcommand{\andor}{AND/OR\xspace}

\title{\lever: Adaptive Cost-Aware Proof Search Over AND/OR Graphs}

\iclrfinalcopy 

\author{%
    Nihal Jain \quad Shuangjie Yao \quad Begum Cicekdag \quad Zhuo Zhang \quad Suman Jana\\
    Columbia University\\
    \texttt{\{nihal.jain, sy3437, bc2975, zz3474, sj2754\}@columbia.edu}
  }

\begin{document}

\maketitle

\begin{abstract}
Mathematicians value proofs for more than correctness: among correct proofs, simplicity, purity and the computational cost of finding them vary widely. Yet LLM-powered theorem provers largely search for \emph{any} correct proof, and improve its quality only after it is found. We propose \lever, a proof search algorithm that makes the objective over correct proofs programmable and optimizes it \emph{during search}. \lever scores partial proofs over an \andor proof graph, combining realized objective values with predictions for open subgoals, so the objective guides search before a proof is complete. The same mechanism optimizes computational cost, proof length, topical impurity, and even their weighted combinations, while the Lean kernel enforces correctness. On PutnamBench in Lean~4, under matched budgets, \lever costs $34\%$ less than a strong single-conversation agent while raising the solve rate from $80\%$ to $96\%$. On reducing topical impurity, \ie, how far a proof strays from its theorem's subject, it improves over post-hoc refactoring ($42\%$ reduction against $33\%$) at two-thirds of the cost and more reliably; on proof length, the metric refactoring is built for, it approaches refactoring. Varying the objective's weights traces a quality--cost trade-off curve, so the user can choose how much a better proof is worth. Overall, \lever is a performant, cost-efficient and tunable proof search algorithm for navigating the space of correct proofs.

\end{abstract}

\section{Introduction} \label{sec:introduction}

Mathematicians have long valued proofs for more than their correctness. A proof may be shorter, more elementary, or reveal structure that another proof obscures. The four-color theorem, for example, was proved in 1976 and formally verified decades later, yet a substantially different proof fifty years on exposed new structure in planar graphs and led to a near-linear-time coloring algorithm~\citep{appel1977every,gonthier2008formal,inoue2026fourcolor}. More broadly, the mathematical literature has debated the importance of characteristics like purity, simplicity and explanation~\citep{steiner1978,detlefsen2011purity,thiele2003hilbert,tao2007good}. The challenge is, therefore, not only to find a correct proof, but to choose among many.

This observation is crucial in the context of LLM-powered theorem provers that may generate correct proofs differing sharply in the computation needed to find them, their length, and the machinery they invoke. Yet current systems largely separate these concerns: proof search returns any verified proof~\citep{nearaiPutnambenchDeepseek,MistralAI_2026,rögnvaldsson2026optimizingcostqualitytradeoffagentic}, sometimes with mechanisms to reduce inference cost, and proof quality is improved only afterwards through post-hoc refactoring~\citep{improver,leanrefactor,fu2026proofrefactorrefactoringgeneratedformal}. We take a different view: \textbf{the objective over the space of proofs should be programmable, with correctness as a constraint}. Optimizing such objectives during search can be both cheaper and more effective than optimizing it afterwards: a post-hoc method revisits a proof already paid for, and possibly inherits a trajectory (mathematical argument) chosen for a different objective.

The key idea of \lever is to score a proof \emph{before} it is finished. A partial proof has some goals proved and others still open. \lever scores it as the realized scored of the proved parts plus a predicted value for each open goal; at any time only the partial proof with the best score is expanded, so the objective steers search from the beginning rather than after the task is complete. We support this mechanism by representing partial proofs as an \andor graph: a theorem or a goal (OR node) is proved by any one of its alternative decompositions (AND nodes), and an argument needs all of its subgoals (OR nodes). We demonstrate that objectives such as computational cost, proof length or topical impurity, or a weighted sum of them compose over the proof graphs, allowing us to optimize them. In this process, correctness is never compromised: the Lean kernel checks every decomposition, and the assembled final proof.

On PutnamBench \citep{tsoukalas2024putnambench}, with the same model and budget allocation as a strong agent that generates proofs in a single conversation, \lever raises the solve rate under a \$3 budget from $80.0\%$ to $96.3\%$ while reducing average cost from \$1.44 to \$0.95, a $34\%$ reduction. The gain comes from keeping contexts local: \lever reads less than half the input tokens of the single-conversation agent. We then optimize the proof itself, for length and \emph{topical impurity}, which we operationalize following work on purity of method \citep{detlefsen2011purity,martinot2024purity}: a proof accumulates impurity when it depends on results outside the subject of its theorem. So, length is often amenable to \textit{local} compression, whereas purity can depend on the argument chosen at the start. Consistent with this, \lever approaches post-hoc refactoring on length, but outperforms both refactoring and direct instruction on impurity, at two-thirds of refactoring's cost. Finally, varying the weight on quality metrics traces explicit quality--cost trade-off curves. This turns theorem proving from searching for \emph{any verified proof} into searching for a verified proof under a user-specified objective, using the \emph{same} search algorithm.

Our contributions are:
\begin{itemize}[leftmargin=*,itemsep=2pt,topsep=2pt]
\item \textbf{Objective-guided search over partial proofs.} A unified algorithm to optimize computation and proof quality during search by using values that combine realized costs with predictions for open subgoals, propagated through an \andor graph; 
\item \textbf{Cheaper and stronger proof search.} On PutnamBench, $34\%$ lower cost and a $96.3\%$ solve rate, against $80.0\%$ for a strong single-conversation agent under a matched budget;
\item \textbf{Proof quality during construction.} Search-time optimization outperforms refactoring on topical impurity at two-thirds of its cost, approaches it on length, and admits explicit quality--cost trade-off curves.
\end{itemize}

\section{Related Work} \label{sec:related-work}

\textbf{LLMs for theorem proving.} LLM provers range from tactic generators inside tree search~\citep{polu2020generative,yang2023leandojo,xin2024deepseekproverv15,xin2025bfsprover} to whole-proof generators refined with compiler feedback~\citep{first2023baldur,wang2025kimina,lin2025goedelproverv2}. Decomposition has since become central: an informal draft becomes a formal sketch whose subgoals are proved separately~\citep{jiang2023draft,wang2024legoprover,dong2024formal,ren2025deepseekproverv2}, and agentic systems such as HILBERT~\citep{hilbert} and Goedel-Architect~\citep{chung2026goedelarchitectstreamliningformaltheorem} complete these, while others learn from millions of attempts~\citep{chen2025seedprover,achim2025aristotleimolevelautomatedtheorem,hubert2025alphaproof,tsoukalas2026alphaproofnexus}; LEAP~\citep{kung2026leap} maintains an \andor graph but explores it depth-first; all these systems search for \emph{any} correct proof. On cost, \citet{rögnvaldsson2026optimizingcostqualitytradeoffagentic} decide when to restart from a new decomposition, and Quarry~\citep{quarry2026} ranks decompositions by predicted hammer solvability; in either case simple linear models with limited features from Lean structures make predictions. Proof quality is improved by refactoring a finished proof~\citep{improver,gu2025proofoptimizertraininglanguagemodels,leanrefactor,fu2026proofrefactorrefactoringgeneratedformal}. \lever instead lets an explicit objective over cost and proof quality steer the search using LLM-based value functions.

  \textbf{Value-guided search.} AO$^*$ finds the cheapest solution graph of an \andor graph under an admissible heuristic~\citep{martelli1973additive,nilsson2014principles}, with extensions to cyclic
  graphs~\citep{hansen2001lao} and games~\citep{allis1994proof}. Learned values replaced hand-written heuristics in Monte Carlo tree
  search~\citep{kocsis2006bandit,silver2016mastering,silver2018general,guez2018learningsearchmctsnets} and HyperTree Proof Search~\citep{lample2022hypertree}; closest in spirit, Retro*~\citep{chen2020retro}
  learns the cost-to-go of \andor trees for retrosynthesis. LLM search over partial solutions~\citep{yao2023tree,hao2023reasoning,zhou2024language,snell2024scaling} is typically guided by self-evaluation or
  process rewards that score correctness~\citep{lightman2023lets,wang2024mathshepherd}. \lever differs in three ways: its expansions are stochastic, so a goal can be redrawn; its priors learn from the draws
  already paid for; and its objective composes the remaining computational cost, the concern of rational metareasoning and costly search~\citep{russell1991principles,desabbata2024rational,weitzman1979optimal},
  with arbitrary proof metrics.

\section{Lever} \label{sec:lever}

\subsection{Background} \label{sec:background}

\begin{wrapfigure}{r}{0.45\textwidth}
  \vspace{-\baselineskip}
  \centering
  \begin{tikzpicture}[
      >={Latex[length=1.7mm, width=1.3mm]},
      goal/.style={circle, draw=ink, line width=0.7pt, minimum size=5mm, inner sep=0pt},
      att/.style={rectangle, fill=ink, draw=ink, minimum size=2.4mm, inner sep=0pt},
      solved/.style={att, double=white, double distance=0.7pt, line width=0.45pt, minimum size=2.2mm},
      prior/.style={rectangle, draw=ink!75, dash pattern=on 1.1pt off 0.8pt, line width=0.6pt, minimum size=2.8mm, inner sep=0pt},
      link/.style={->, draw=ink, line width=0.6pt, shorten >=0.4pt},
      plink/.style={->, draw=ink!75, line width=0.7pt, densely dotted},
      lab/.style={font=\footnotesize, text=ink, inner sep=1pt},
      sub/.style={lab, text height=1.7ex, text depth=0.6ex},
      tag/.style={font=\footnotesize\itshape, text=muted, inner sep=1pt},
    ]
    \def\ya{0}\def\yb{-1.3}\def\yc{-2.6}
    \node[goal] (r) at (0, \ya) {};
    \node[lab, anchor=west] at ([xshift=0.8mm]r.east) {infinitely many primes};
    \node[prior] (p) at (-1.5, \ya) {};
    \draw[plink] (r) -- (p);
    \node[tag, anchor=east, text=ink!75] at ([xshift=-0.8mm]p.west) {prior};
    \node[att] (a1) at (-1.3, \yb) {};
    \node[att] (a2) at (1.6, \yb) {};
    \draw[link] (r) -- (a1);
    \draw[link] (r) -- (a2);
    \node[tag, anchor=east, text=hicyan] at ([xshift=-0.8mm]a1.west) {Euclid};
      \node[tag, anchor=west, text=amber!85!black] at ([xshift=0.8mm]a2.east) {Fermat};
    \node[goal] (g1) at (-2.5, \yc) {};
    \node[goal] (g2) at (-1.3, \yc) {};
    \node[goal] (g3) at (-0.1, \yc) {};
    \draw[link] (a1) -- (g1);
    \draw[link] (a1) -- (g2);
    \draw[link] (a1) -- (g3);
    \node[sub, anchor=north] (t1) at ([yshift=-0.3mm]g1.south) {$N{>}1$};
    \node[sub, anchor=north] (t2) at ([yshift=-0.3mm]g2.south) {$p \mid N$};
    \node[sub, anchor=north] (t3) at ([yshift=-0.3mm]g3.south) {$p \neq p_i$};
    \node[goal] (g4) at (1.6, \yc) {};
    \draw[link] (a2) -- (g4);
    \node[sub, anchor=north] (t4) at ([yshift=-0.3mm]g4.south) {$(F_m, F_n) = 1$};
    \foreach \t in {t1, t2} {
      \draw[link] (\t.south) -- ++(0, -5mm) coordinate (tip);
      \node[solved, anchor=north] at (tip) {};
    }
    \foreach \t in {t3, t4} {
      \draw[link] (\t.south) -- ++(0, -5mm) coordinate (tip);
      \foreach \k in {1, 2, 3}
        \fill[ink] ([yshift=-\k*1.1mm-0.3mm]tip) circle (0.4pt);
    }
  \end{tikzpicture}
  \caption{\textbf{An AND/OR graph.} OR nodes (goals) are circles, AND nodes (decompositions) are squares; a goal may have different decompositions (here Euclid's and Fermat's). Double border AND nodes are proved directly; dashed AND nodes are priors~(\S\ref{sec:adaptive-proof-search-with-lever}).}
  \label{fig:andor-small}
  \vspace{-\baselineskip}
\end{wrapfigure}
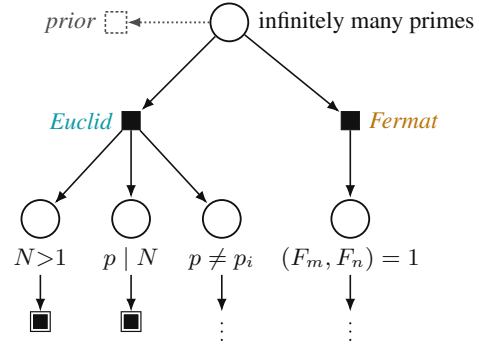

\textbf{AND/OR graphs.} An AND/OR graph is a bipartite DAG $G=(V_{OR} \sqcup V_{AND}, E)$ whose edges alternate between OR and AND nodes. An OR node is a \textit{goal}, a theorem or lemma to prove; an AND node is a \textit{decomposition} of a goal into \textit{subgoals}, whose validity an oracle (here the Lean kernel) checks. The root OR node $r$ is the theorem to prove. An OR node is solved if \textit{any} child is, and an AND node if \textit{all} its children are; an AND node with no children, a goal proved without new lemmas, is solved.

\textbf{Solution subgraphs.} A \textit{proof} or \textit{solution subgraph} is a subgraph $S\subseteq G$ such that: (i) $r\in S$; (ii) for each OR node in $S$, exactly one AND child is picked; (iii) for every AND node in $S$, all its children are picked; (iv) the leaves in $S$ are solved. Then each $S$ maps to a unique proof written in Lean.

\textbf{Costs.} Every edge carries a nonnegative cost $c$. A decomposition needs to pay for all its subgoals whereas a goal takes the cheapest way to prove it, so values back up as:

$$V(n)=\sum_{d\in\mathrm{ch}(n)} c(n,d)+V(d) \quad (n\in V_{AND}), \qquad V(n)=\min_{d\in\mathrm{ch}(n)} c(n,d)+V(d) \quad (n\in V_{OR}),$$

with $V=0$ at a solved leaf and $V=\infty$ at a false goal. $V(r)$ is then the cost of the cheapest solution subgraph, and the argmin at each OR node reconstructs it.

\subsection{Proof Search} \label{sec:proof-search}

The problem of proof search can be modeled as an MDP with the following description:
\begin{itemize}[leftmargin=*,itemsep=2pt,topsep=2pt]
    \item \textbf{States.} The graph built so far, with statuses and edge costs; fully observed.
    \item \textbf{Actions.} OR node(s) to expand, \ie, which goals to decompose next or STOP if nothing to expand.
    \item \textbf{Transition.} Transformation of the graph that adds decompositions (and the corresponding subgoals) to the graph; transitions are stochastic: expanding the same goal twice may not result in the same resulting graph.
    \item \textbf{Reward.} Score of $-s_t$ for the expansions at time step $t$, and $-V(r)$ at termination ($-\infty$ if unsolved). Episodes end when the policy decides to STOP or if the budget is exhausted.
\end{itemize}
The objective of proof search is to maximize the sum of rewards at the end of the episode or minimize the cost objective: $V(r) + \sum_{t} s_t$. At any step, the spend so far is sunk and cannot be changed, so a good policy minimizes the \emph{cost-to-go}: the expected \emph{remaining} spend plus $V(r)$.

\subsection{Adaptive Proof Search with \lever} \label{sec:adaptive-proof-search-with-lever}

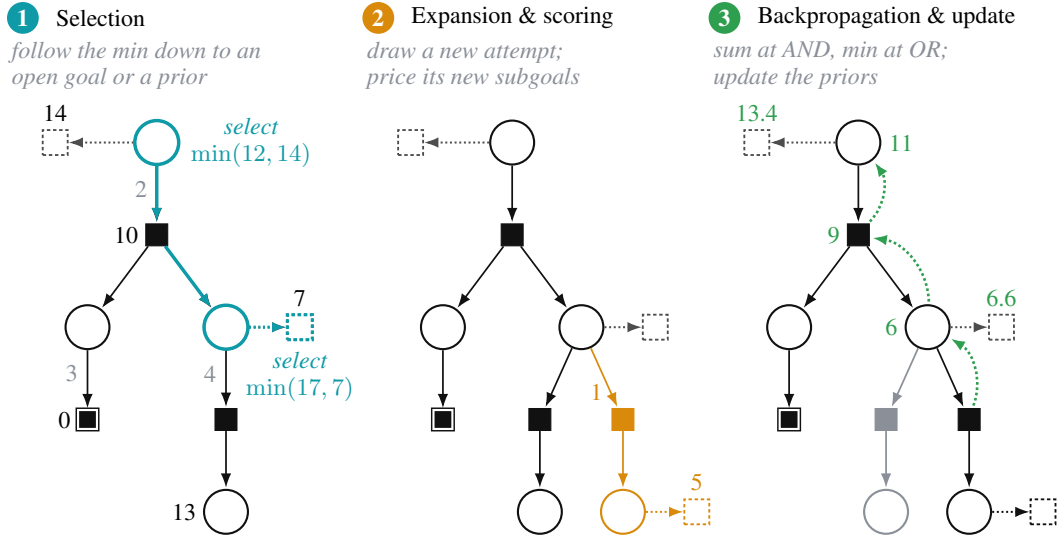
\begin{figure}[t]
  \centering
  \resizebox{\textwidth}{!}{
  \begin{tikzpicture}[
      scale=1.22,
      >={Latex[length=1.9mm, width=1.4mm]},
      goal/.style={circle, draw=ink, line width=0.8pt, minimum size=5.8mm, inner sep=0pt},
      att/.style={rectangle, fill=ink, draw=ink, minimum size=2.9mm, inner sep=0pt},
      solved/.style={att, double=white, double distance=0.8pt, line width=0.55pt, minimum size=2.5mm},
      prior/.style={rectangle, draw=ink!75, dash pattern=on 1.3pt off 1pt, line width=0.7pt, minimum size=3.3mm, inner sep=0pt},
      link/.style={->, draw=ink, line width=0.65pt, shorten >=0.4pt},
      plink/.style={->, draw=ink!75, line width=0.75pt, densely dotted},
      sel/.style={draw=hicyan, line width=1.3pt},
      back/.style={->, draw=upgreen, line width=1pt, densely dotted, shorten >=1pt, shorten <=1pt},
      val/.style={font=\small, inner sep=0.7pt},
      cost/.style={font=\small, inner sep=0.7pt, text=muted},
      pick/.style={font=\footnotesize\itshape, text=hicyan, align=center, inner sep=0.5pt},
      badge/.style={circle, fill=#1, text=white, font=\bfseries\footnotesize, inner sep=0pt, minimum size=4.2mm},
      steptitle/.style={font=\small, anchor=west},
      stepnote/.style={font=\footnotesize\itshape, text=muted, anchor=north west, align=left, inner sep=0pt},
    ]
    \newcommand{\basegraph}[3]{%
      \node[goal] (r#1) at (0, 0) {};
      \node[prior] (pr#1) at (-1.1, 0) {};
      \node[att] (a1#1) at (0, -1.0) {};
      \node[goal] (g1#1) at (-0.75, -2.0) {};
      \node[goal] (g2#1) at (0.75, -2.0) {};
      \node[solved] (l1#1) at (-0.75, -3.0) {};
      \node[prior] (p2#1) at (1.55, -2.0) {};
      \node[att, fill=#2, draw=#2] (a2#1) at (#3, -3.0) {};
      \node[goal, draw=#2] (g3#1) at (#3, -4.0) {};
      \draw[plink] (r#1) -- (pr#1);
      \draw[link] (r#1) -- (a1#1);
      \draw[link] (a1#1) -- (g1#1);
      \draw[link] (a1#1) -- (g2#1);
      \draw[link] (g1#1) -- (l1#1);
      \draw[plink] (g2#1) -- (p2#1);
      \draw[link, draw=#2] (g2#1) -- (a2#1);
      \draw[link, draw=#2] (a2#1) -- (g3#1);
    }
    \newcommand{\newattempt}[2]{
      \node[att, fill=#2, draw=#2] (a4#1) at (1.2, -3.0) {};
      \node[goal, draw=#2] (g4#1) at (1.2, -4.0) {};
      \node[prior, draw=#2] (p4#1) at (2.0, -4.0) {};
      \draw[link, draw=#2] (g2#1) -- (a4#1);
      \draw[link, draw=#2] (a4#1) -- (g4#1);
      \draw[plink, draw=#2] (g4#1) -- (p4#1);
    }

    \begin{scope}
      \basegraph{A}{ink}{0.75}
      \node[cost, anchor=east] at ($(rA)!0.5!(a1A)+(-0.08,0)$) {2};
      \node[cost, anchor=east] at ($(g1A)!0.5!(l1A)+(-0.08,0)$) {3};
      \node[cost, anchor=east] at ($(g2A)!0.5!(a2A)+(-0.08,0)$) {4};
      \node[val, anchor=south] at ([yshift=0.6mm]prA.north) {14};
      \node[val, anchor=east] at ([xshift=-0.5mm]a1A.west) {10};
      \node[val, anchor=east] at ([xshift=-0.5mm]l1A.west) {0};
      \node[val, anchor=south] at ([yshift=0.6mm]p2A.north) {7};
      \node[val, anchor=east] at ([xshift=-0.5mm]g3A.west) {13};
      \node[goal, sel] at (rA) {};
      \node[goal, sel] at (g2A) {};
      \node[prior, draw=hicyan, line width=1.2pt] at (p2A) {};
      \draw[link, sel] (rA) -- (a1A);
      \draw[link, sel] (a1A) -- (g2A);
      \draw[plink, draw=hicyan, line width=1.2pt] (g2A) -- (p2A);
      \node[pick, anchor=west] at ([xshift=1mm]rA.east) {select\\$\min(12, 14)$};
      \node[pick, anchor=north] at ([yshift=-1mm]p2A.south) {select\\$\min(17, 7)$};
      \node[badge=hicyan] (b1) at (-1.45, 1.35) {1};
      \node[steptitle] at ([xshift=0.8mm]b1.east) {Selection};
      \node[stepnote] at ([yshift=-1.2mm]b1.south west) {follow the min down to an\\open goal or a prior};
    \end{scope}

    \begin{scope}[xshift=3.85cm]
      \basegraph{B}{ink}{0.3}
      \newattempt{B}{amber}
      \node[cost, anchor=east, text=amber] at ($(g2B)!0.7!(a4B)+(-0.06,0)$) {1};
      \node[val, anchor=south, text=amber] at ([yshift=0.6mm]p4B.north) {5};
      \node[badge=amber] (b2) at (-1.45, 1.35) {2};
      \node[steptitle] at ([xshift=0.8mm]b2.east) {Expansion \& scoring};
      \node[stepnote] at ([yshift=-1.2mm]b2.south west) {draw a new attempt;\\price its new subgoals};
    \end{scope}

    \begin{scope}[xshift=7.6cm]
      \basegraph{C}{muted}{0.3}
      \newattempt{C}{ink}
      \draw[back] (a4C) to[bend right=40] (g2C);
      \draw[back] (g2C) to[bend right=40] (a1C);
      \draw[back] (a1C) to[bend right=40] (rC);
      \node[val, text=upgreen, anchor=west] at ([xshift=0.6mm]rC.east) {11};
      \node[val, text=upgreen, anchor=south] at ([yshift=0.6mm]prC.north) {13.4};
      \node[val, text=upgreen, anchor=east] at ([xshift=-0.5mm]a1C.west) {9};
      \node[val, text=upgreen, anchor=east] at ([xshift=-0.5mm]g2C.west) {6};
      \node[val, text=upgreen, anchor=south] at ([yshift=0.6mm]p2C.north) {6.6};
      \node[badge=upgreen] (b3) at (-1.45, 1.35) {3};
      \node[steptitle] at ([xshift=0.8mm]b3.east) {Backpropagation \& update};
      \node[stepnote] at ([yshift=-1.2mm]b3.south west) {sum at AND, min at OR;\\update the priors};
    \end{scope}
  \end{tikzpicture}}
  \caption{\textbf{One iteration of \lever.} Grey numbers are edge costs and black numbers node values; later panels label only what their step changes. At the root's second subgoal the prior beats the realized attempt,
  so that interior goal is expanded again; the new attempt's value then passes up to the root, updating the priors on the way (Eq.~\ref{eq:conjugate}, $n = 1$). The first attempt stays in the graph, paid for but no
  longer chosen (grey).}
  \label{fig:lever-steps}
\end{figure}

Algorithm~\ref{alg:lever} shows how \lever searches for proofs. The AO$^*$ algorithm~\citep{nilsson2014principles} assumes a deterministic expansion function, but here the expander is an LLM: expanding a goal returns one decomposition drawn from an effectively unbounded set, and drawing again may return a cheaper one. Hence a goal that already has children can still be worth expanding, \ie, any \textit{interior} OR node is a candidate. To decide when, every OR node $g$ carries one additional child, its \textit{prior}: an AND node $p$ that is never expanded, holding an estimate $\tilde{V}(g)$ of the cost of finishing $g$ with a fresh decomposition. Each iteration has three steps (Figure~\ref{fig:lever-steps}):

\begin{algorithm}[t]
  \DontPrintSemicolon
  \caption{\lever. Each iteration selects goals along the current best solution subgraph, expands them, attaches a prior to each new subgoal, and backpropagates values, until selection returns nothing or the budget is spent.} \label{alg:lever}
  \SetKwFunction{FSelect}{Select}
  \SetKwFunction{FPrior}{Prior}
  \SetKwFunction{FExpand}{Expand}
  \SetKwFunction{FBackup}{Backpropagate}
  \KwIn{root $r$, budget $B$}
  \FPrior{$r$}\;
  \While{$\mathrm{spent} < B$}{
    $T \leftarrow$ \FSelect{$r$}\;
    \lIf{$T = \emptyset$}{\KwSty{break}}
    \ForEach{$g \in T$}{
      $D \leftarrow$ \FExpand{$g$}\;
      \lForEach{$d \in D$}{\FPrior{$d$}}
      \FBackup{$g$}\;
    }
  }
  \Return{the best solution subgraph of $r$}\;
\end{algorithm}

\textbf{1. Selection.} Descend from the root along the current best solution subgraph, taking the $\argmin$ child at an OR node and all children at an AND node. When the argmin is a prior, its goal is expanded; for a goal that already has children, this is a \emph{redraw}. Values are noisy, so a small gap is no evidence that a fresh draw would be cheaper: a redraw must beat the incumbent by a factor $\delta$. When selection returns nothing, the policy STOPs.

\textbf{2. Expansion.} An LLM expands each selected goal: it proposes a new decomposition (an AND node), the oracle checks it, and each subgoal it introduces is attached as a new OR node. The edge from the goal to the decomposition is scored with the cost of the expansion, and each new subgoal receives a prior from the value function (\S\ref{sec:value-functions}). Edges from a decomposition to its subgoals cost nothing: the subgoals are simply parsed from the compiled proof.

\textbf{3. Backpropagation.} New costs pass upward through the recursion of \S\ref{sec:background}, updating every prior on the path. We model the log cost of a fresh decomposition of $g$ as normal around an unknown mean $\mu_g$ with spread $\sigma$ (draws of the same goal differ) and the value function's prediction $V_0(g)$ (\S\ref{sec:value-functions}) as off from $\mu_g$ by $\rho$. After $k$ decompositions with costs $v_1, \dots, v_k$, the posterior $\mu_g$ is normal with:
\begin{equation}
  \hat\mu_g \;=\; \frac{n \log V_0(g) + \sum_{i=1}^{k} \log v_i}{n + k},
  \qquad n = \sigma^2 / \rho^2,
  \label{eq:conjugate}
\end{equation}
so the prior estimate is worth $n$ draws and each realized draw, one. The value estimate is the posterior geometric mean, $\tilde V(g) = \exp(\hat\mu_g)$. Realized costs thus correct the value function's errors: this is how the search learns from what it has paid for. A failed attempt leaves no cost to learn from, and instead raises the estimate by one $\sigma$. Metrics on a linear scale use the same form without logarithm.

\textbf{Objective.} The steps above hold for any cost that composes over the algebra of \S\ref{sec:background}, which metrics such as proof length do. The spend term $\sum_t s_t$ in the MDP objective of \S\ref{sec:proof-search} does not: it accumulates over the whole search, including draws the final proof never uses. We therefore split it into what has been spent (sunk; cannot be optimized) and what remains (can be optimized). The remaining part does compose over the graph, and we call it $V_{\texttt{comp}}(n)$: the computational cost still needed to finish $n$, where realized edges add nothing and open goals add their prior's estimate. With this adjustment, any metric $m$, backed up as $V_m$ with edge costs $c_m$, can be balanced against computational cost at a price of $\lambda_m$ dollars per unit: $C(n) \;=\; V_{\texttt{comp}}(n) + \sum_m \lambda_m V_m(n).$ With every $\lambda_m = 0$, \lever minimizes computational cost alone.

\textbf{Exactness under exact priors.} With exact priors, the backed-up values are exact: for any nonnegative cost that adds along edges, $V(n)$ is the cost of the cheapest solution subgraph rooted at $n$ (counting a subgoal once per use), and $V(n) = \infty$ exactly when none exists. We also share a formally verified proof of this in Lean (Appendix~\ref{appdx:lemma}). Real value functions are noisy, which the update of Equation~\ref{eq:conjugate} and the margin $\delta$ account for; their theoretical analysis is left for future work. When a subgoal is shared by several decompositions, paying for it once instead makes the cheapest solution subgraph NP-hard to find in general~\citep{sahni1974computationally}; \lever instead charges each shared subgoal to a single owner (Appendix~\ref{appdx:system-design}).

\section{Implementing \lever with Lean} \label{sec:implementation}

\subsection{Setup} \label{sec:implementation-setup}

\textbf{Model and Pricing.} We use DeepSeek-V4-Flash-0731~\citep{deepseekv42026} for every LLM call, in \lever and in all baselines, to match the baseline's original setting. We price tokens at the official rates of \$0.22, \$0.007 and \$0.66 per million input, cached input and output tokens~\citep{deepseekv4priceupdate}. Cost therefore depends heavily on the cache-hit rate. Served through multiple providers, API calls experience mean hit rate per call at 93.7\% for \citet{nearaiPutnambenchDeepseek}, whose single long conversation re-reads its own history, and 86.2\% for \lever, which starts many shorter conversations. To compare the searches rather than the caching, we fixed one 90\% rate, between the two, before the evaluation runs. All costs in this paper are re-priced at this rate, and a problem counts as solved within a budget $B$ only if its re-priced cost is at most $B$; Appendix~\ref{appdx:realized-cost} reports the observed costs.

\textbf{Data.} We use PutnamBench~\citep{tsoukalas2024putnambench}, originally containing $672$ competition problems formalized in Lean~4 (v4.31.0). Most are easy for a strong agent, while the hardest can take days of repeated attempts, too long for controlled comparisons. Hence, we rank the problems by what the published run of \citet{nearaiPutnambenchDeepseek} spent to solve them and take ranks $51$--$150$, a slice that is hard but tractable; Appendix~\ref{appdx:data} gives the full procedure and its biases. $20$ of these calibrate the value functions (\S\ref{sec:value-functions}), and the other $80$ form the evaluation set (\S~\ref{sec:experiments}).

\subsection{Agentic System Design} \label{sec:system-design}

We discuss salient aspects of \lever's agent for goal expansion, leaving detailed descriptions in Appendix~\ref{appdx:system-design}.

\textbf{Expanding a goal.} Each expansion is one conversation of an LLM agent~\citep{yao2023react} with a Lean workspace, with tools for Lean compilation~\citep{santos2025kiminaleanserverhighperformance}, Mathlib retrieval~\citep{leanfinder}, Python, and file I/O; it ends by submitting a file or reporting the goal false. To \emph{decompose} a goal, the agent writes a Lean proof that relies on open subgoals; small lemmas are proved in place and larger ones are left open. Each open helper becomes a subgoal, to be closed in a future iteration. It annotates dependencies among open subgoals with the \texttt{sorry\_using} tactic of LeanArchitect~\citep{leanarchitect,chung2026goedelarchitectstreamliningformaltheorem}, so that subgoals can share work. To prove a goal \emph{directly}, the agent writes a complete proof with no open subgoals; \S\ref{sec:experiments} states when this is enforced.

\textbf{Context.} Before the search, a planner writes an informal proof of the problem, following \citet{hilbert} and \citet{chung2026goedelarchitectstreamliningformaltheorem}. Every expansion carries this plan in its system prompt, and the planner's cost is included in every cost we report for \lever. A subgoal's conversation starts from its parent's compacted output, or its full transcript if short, so the agent knows why the subgoal was introduced without carrying its parent's work (Appendix~\ref{appdx:system-design}). Every LLM inference call is made with $32$K completion tokens at temperature $1.0$; planning with high-effort reasoning and expansion with low-effort reasoning.

\textbf{Disproofs and soundness.} The agent may report a goal false, with an informal reason (formal disproofs as in \citet{lin2025goedelproverv2} are not integrated for simplicity; these can wander into their own long trajectories). This ends the decomposition, and values are updated as in \S\ref{sec:adaptive-proof-search-with-lever}. When the root is solved, the solution subgraph is assembled into one Lean file and checked by the kernel. A problem counts as solved only if this file compiles using only the standard axioms, without \texttt{sorry}, \texttt{admit}, \texttt{axiom} or \texttt{native\_decide}.

 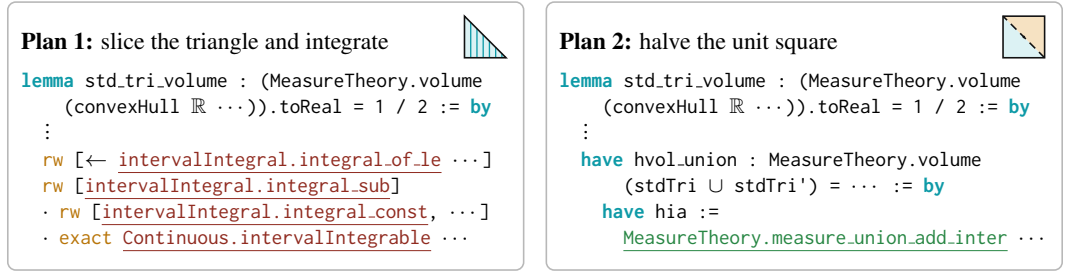
\begin{figure}[t]
  \centering
\lstdefinelanguage{lean4}{
  morekeywords=[1]{lemma, theorem, def, noncomputable, have, by, fun, let, in, at, with, calc, show},
  morekeywords=[2]{rw, simp, simpa, exact, norm_num, linarith, nlinarith, refine, intro, rintro, apply, ext, unfold, using},
  morecomment=[l]{--},
  alsoletter={.,_,'},
  sensitive=true,
}
\lstdefinestyle{leanfig}{
  language=lean4,
  basicstyle=\fontsize{8}{9.8}\fontfamily{zi4}\selectfont,
  keywordstyle=[1]\color{hicyan}\bfseries,
  keywordstyle=[2]\color{amber!85!black},
  commentstyle=\color{muted}\itshape,
  escapeinside={(*}{*)},
  columns=fullflexible, keepspaces=true, breaklines=false, upquote=true,
  xleftmargin=0pt, aboveskip=3pt, belowskip=2pt,
  extendedchars=true,
  literate={ℝ}{{$\mathbb{R}$}}1 {∫}{{$\int$}}1 {∂}{{$\partial$}}1
           {≤}{{$\le$}}1 {←}{{$\leftarrow$}}1 {·}{{$\cdot$}}1
           {⋯}{{$\cdots$}}1 {∪}{{$\cup$}}1 {×}{{$\times$}}1
           {⋮}{{\vbox{\baselineskip=2.6pt\lineskiplimit=-9pt\hbox{.}\hbox{.}\hbox{.}}}}1,
}
\providecommand{\flag}[1]{\textcolor{termred!75!black}{\smash[b]{\underline{#1}}}}
\providecommand{\pure}[1]{\textcolor{upgreen!85!black}{\smash[b]{\underline{#1}}}}
\providecommand{\leanname}[1]{{\fontfamily{zi4}\selectfont #1}}
\newsavebox{\purityplan}
\newcommand{\planframe}{\tikz[baseline=(f.north)]
  \node[draw=ink!30, line width=0.6pt, rounded corners=3pt, inner sep=5pt] (f) {\usebox{\purityplan}};}
\newcommand{\planwidth}{\dimexpr0.49\textwidth-10.6pt\relax}
\begin{lrbox}{\purityplan}\begin{minipage}[t]{\planwidth}
  \begin{tikzpicture}[baseline=(t.base)]
    \node[anchor=base west, inner sep=0pt] (t) at (0, 0) {\small\textbf{Plan 1:} slice the triangle and integrate};
    \begin{scope}[shift={(\linewidth - 0.6cm, -0.12)}, scale=0.55]
      \fill[hicyanfill] (0, 0) -- (1, 0) -- (0, 1) -- cycle;
      \foreach \a in {0.15, 0.3, 0.45, 0.6, 0.75, 0.9}
        \draw[hicyan, line width=0.5pt] (\a, 0) -- (\a, 1 - \a);
      \draw[ink, line width=0.6pt] (0, 0) -- (1, 0) -- (0, 1) -- cycle;
    \end{scope}
  \end{tikzpicture}
\begin{lstlisting}[style=leanfig]
lemma std_tri_volume : (MeasureTheory.volume
    (convexHull ℝ ⋯)).toReal = 1 / 2 := by
  ⋮
  rw [← (*\flag{intervalIntegral.integral\_of\_le}*) ⋯]
  rw [(*\flag{intervalIntegral.integral\_sub}*)]
  · rw [(*\flag{intervalIntegral.integral\_const}*), ⋯]
  · exact (*\flag{Continuous.intervalIntegrable}*) ⋯
\end{lstlisting}
\end{minipage}\end{lrbox}\planframe\hfill
\begin{lrbox}{\purityplan}\begin{minipage}[t]{\planwidth}
  \begin{tikzpicture}[baseline=(t.base)]
    \node[anchor=base west, inner sep=0pt] (t) at (0, 0) {\small\textbf{Plan 2:} halve the unit square};
    \begin{scope}[shift={(\linewidth - 0.6cm, -0.12)}, scale=0.55]
      \fill[hicyanfill] (0, 0) -- (1, 0) -- (0, 1) -- cycle;
      \fill[amber!25] (1, 1) -- (0, 1) -- (1, 0) -- cycle;
      \draw[ink, line width=0.6pt] (0, 0) rectangle (1, 1);
      \draw[ink, line width=0.6pt, densely dashed] (1, 0) -- (0, 1);
      \fill[ink] (0.5, 0.5) circle (0.9pt);
    \end{scope}
  \end{tikzpicture}
\begin{lstlisting}[style=leanfig]
lemma std_tri_volume : (MeasureTheory.volume
    (convexHull ℝ ⋯)).toReal = 1 / 2 := by
  ⋮
  have hvol_union : MeasureTheory.volume
      (stdTri ∪ stdTri') = ⋯ := by
    have hia :=
      (*\pure{MeasureTheory.measure\_union\_add\_inter}*) ⋯
\end{lstlisting}
\end{minipage}\end{lrbox}\planframe
  \caption{\textbf{Two proofs of one lemma: the unit triangle has area $\tfrac12$.} The statement is about measure and convex hulls. Plan 1 slices the triangle and integrates. Its calculus references, such as
  \flag{\leanname{intervalIntegral.integral\_sub}}, are impure (underlined red): the statement never calls for integrals. Plan 2 halves the unit square. It argues in measure theory, and a reference such as
  \pure{\leanname{MeasureTheory.measure\_union\_add\_inter}} is pure (underlined green). Over the lemma and the helpers it uses, the metric flags 25 references in Plan 1's proof and 6 in Plan 2's. Excerpts from compiled
  proofs; dots mark elisions.}
  \label{fig:purity-example}
  \end{figure}

\subsection{Cost Objectives} \label{sec:cost-objectives}

\textbf{Correctness and computational cost.} Correctness is enforced as a constraint: an unsolved goal costs $\infty$, so every objective is minimized over proofs that Lean type-checks. Computational cost, \texttt{comp}, is the dollars spent on LLM calls at the rates of \S\ref{sec:implementation-setup}. Spend already incurred cannot be recovered, so the search never optimizes it: each decision minimizes the \emph{remaining} cost of finishing the proof, as estimated by the value function (\S\ref{sec:value-functions}), and every expansion replaces part of that estimate with what it actually cost (\S\ref{sec:adaptive-proof-search-with-lever}).

\textbf{Proof length.} We count the Lean tokens (\texttt{len}) a proof adds to its theorem statement, helper lemmas and definitions included and comments excluded, tokenized as in \citet{gu2025proofoptimizertraininglanguagemodels}. It is easy to measure but also easy to game, \eg, a goal closed by one call to \texttt{nlinarith} is short while saying nothing about the mathematics behind it.

\textbf{Topical impurity.} (\texttt{imp}) instead measures the source of content in a proof. A proof is \emph{pure} when it uses only what its theorem is about~\citep{detlefsen2011purity}. Purity is not computable in general, since a human must decide what a theorem's content is~\citep{martinot2024purity}, so we read the content off Mathlib's structure~\citep{The_mathlib_Community_2020}. Every library name has a namespace and a directory, \eg, \texttt{ZMod.wilsons\_lemma} is about \texttt{ZMod} and lives in \texttt{Mathlib/NumberTheory}. A theorem's \emph{subject} is the namespaces and directories of the names its statement uses. A referenced library theorem is \emph{impure} when it falls outside the subject on both accounts; a generic lemma with no namespace, such as \texttt{mul\_pos}, is impure only if it also lies outside a fixed set of infrastructure directories (\eg, \texttt{Mathlib/Data/Nat}) that proofs of every topic use. Overall, we measure impurity as the number of impure references. See Figure~\ref{fig:purity-example} for an illustration of pure and impure proofs. We note here that this is a proxy for measuring mathematical (im)purity; further discussion and limitations of this metric are in Appendix~\ref{appdx:topical-impurity}.

 \subsection{Value Functions} \label{sec:value-functions}

\textbf{Predicting \texttt{comp} and \texttt{len}.} For \texttt{comp} and \texttt{len}, an LLM reads the open goal and picks one of seven bins on a $\log_2$ scale (\$0.016--\$0.512 and 512--16{,}384 Lean tokens), and the goal is priced at its bin's geometric center. Further, for \texttt{len}, the estimate is $0.2$ times the bin's center, deliberately underestimating length so that shorter proofs get explored. The prompt carries $40$ in-context examples from offline runs on the calibration problems. Using an LLM as a value function prior is inexpensive: these calls make up only $0.89\%$ of \lever's spend, and are always included in the reported costs.

\textbf{Predicting impurity.} Impurity has a real floor, since a proof can cite nothing outside its subject, so the natural estimate for an open goal is $0$. In practice this made search unstable: every open goal looked pure, so redrawing always beat keeping a decomposition, and graphs grew huge. We instead score every open goal at a constant $4$ references, the $25$th percentile of impurity over calibration goals.

 \textbf{Calibration.} On the calibration problems only, we measure $\rho$, the value function's error against realized values, and $\sigma$, the spread of a goal's realized value across repeated draws: in log space for \texttt{comp} and \texttt{len}, and in references for \texttt{imp}. For \texttt{comp} and \texttt{len}, the redraw margin is one spread, $\delta = e^{\sigma}$. Appendix~\ref{appdx:value-functions} documents every constant and how it is measured.

\section{Experiments} \label{sec:experiments}

We focus on two main questions: First, can \lever optimize computational costs during proof search~(\S~\ref{sec:optimizing-computational-cost})? Second, how effective is \lever at recovering high quality proofs during search~(\S~\ref{sec:proof-quality-optimization})? All runs share the setup of~\S~\ref{sec:implementation-setup}: one model, one Lean version and one price table, for \lever and every baseline alike.

\subsection{Optimizing Computational Cost} \label{sec:optimizing-computational-cost}

\textbf{Baseline.} Our main baseline is the open-source agent of \citet{nearaiPutnambenchDeepseek}, the top entry on the PutnamBench leaderboard at the time of writing. It is a single-conversation agent: one conversation holds the whole proof and every compiler message. We run the released agent with its documented 1M-token context and compaction, under the same model, prices and Lean version as \lever (\S\ref{sec:implementation-setup}).

\textbf{Setup.} We compare \lever with this agent. \lever uses the \texttt{comp} value function (\S\ref{sec:value-functions}) during decomposition. In our exploration, we found that the LLM used here tends to over-decompose, even on simple problems. So, a goal five levels below the root is always proved directly, and \lever uses its value function to force a direct proof for any goal priced under bin 5 (\S\ref{sec:value-functions}). Each problem has a budget of \$3 at the re-priced rate (\S\ref{sec:implementation-setup}), and a problem not solved within it is charged the full \$3. We evaluate on the $80$ evaluation problems of \S\ref{sec:implementation-setup}.

\begin{figure}[ht]
    \centering
    \begin{minipage}[c]{0.57\textwidth}
        \centering
        \setlength{\tabcolsep}{4pt}
        \begin{tabular}{l c c r r}
      \toprule
      & & & \multicolumn{2}{c}{Tokens (M)} \\
      \cmidrule(lr){4-5}
      Method & Solved & Cost~(\$) & In & Out \\
      \midrule
        NearAI     & 0.800 & 1.44 & 42.0 & \textbf{0.36} \\
        \lever     & \textbf{0.963} & \textbf{0.95} & \textbf{17.9} & 0.62 \\
      \bottomrule
  \end{tabular}
     \captionof{table}{\textbf{Cost of proof search at a \$3 budget per problem.} $N=80$. Cost and tokens are per problem. Solved problems contribute the cost to finish; unsolved problems are charged the full budget with tokens spent up to the call when cut-off.} \label{tab:cost-results}
    \end{minipage}\hfill
    \begin{minipage}[c]{0.41\textwidth}
        \centering
        \includegraphics[width=\linewidth]{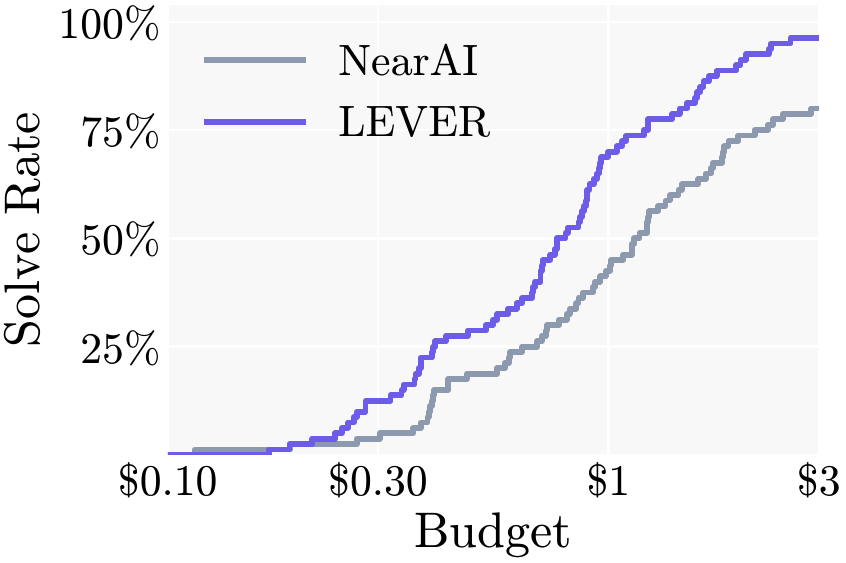}
        \captionof{figure}{\textbf{Solve rate against budget.} A problem solved at cost $c$ counts as solved at every budget $\geq c$.}
        \label{fig:solve-vs-budget}
    \end{minipage}
\end{figure}

\textbf{Decomposition makes proof search cheaper.} Table~\ref{tab:cost-results} compares the two systems at \$3. \lever solves $96.3\%$ of the problems, against $80.0\%$ for NearAI, and costs $34\%$ less. This is not a trade-off between solving and saving: \lever does better on \textit{both}. Further, the failure charge is real spend: $94\%$ of NearAI's failures spent the full \$3 without producing a complete proof. Figure~\ref{fig:solve-vs-budget} shows that this holds at almost every budget: from \$0.25 on, \lever solves more problems than NearAI. The savings come from how \lever manages context. NearAI writes fewer tokens than \lever ($357$K against $621$K per problem), but it carries the context of the \emph{full} problem on every turn, and reads $42.0$M input tokens per problem. \lever works on each subgoal in its own local context and reads $17.9$M, less than half. At the cost each run actually incurred, with its own cache hits, \lever still solves more problems than NearAI ($0.963$ vs. $0.863$) and costs $16\%$ less (Appendix~\ref{appdx:realized-cost}).

\textbf{Identifying cheaper strategies.} \lever's algorithm uses the value function to \emph{reject} decompositions whose subgoals sum up to be more expensive than the goal itself, and draw a new decomposition. On \texttt{putnam\_1972\_a3}, for example, \lever rejects its first decomposition from the value predictions alone, before working on any of its subgoals, and proves the theorem through a second decomposition predicted to be cheaper (Appendix~\ref{appdx:case-study-1972-a3}). However, such cases are rare, and not every redraw leads to savings: every subgraph that is given up on represents a sunk cost that is deducted from the global budget.

\subsection{Proof Quality Optimization} \label{sec:proof-quality-optimization}

 \textbf{Setup.} We optimize the two metrics of \S\ref{sec:cost-objectives}, topical impurity and proof length, on $N=20$ problems drawn from successful runs of NearAI in \S\ref{sec:optimizing-computational-cost}. We write \textsc{Lever}$_m^{\lambda}$ for \lever minimizing the objective $C = \texttt{comp} + \lambda \cdot m$ with $m \in \{\texttt{imp}, \texttt{len}\}$. Further, \textsc{Lever}$_m^0$ is always written as \lever (\ie, the run from \S~\ref{sec:optimizing-computational-cost}). Unlike \lever, which enforces no decomposition under a certain value, \textsc{Lever}$_m^{\lambda}$ runs impose no such restriction to allow greater exploration during search. We run each metric $m$ at two weights, and each run (including all baselines; see below) is with budget \$4.5 (50\% higher than in \S~\ref{sec:optimizing-computational-cost} to accommodate cost of quality optimization on top of correctness).

\textbf{Baselines.} Existing proof refactoring methods that target length, such as Lean Refactor~\citep{leanrefactor} and ProofOptimizer~\citep{gu2025proofoptimizertraininglanguagemodels} are not publicly available, while ImProver~\citep{improver} targets a much older Lean version (v4.9) and model; none can be run in our controlled setup, with the same model, Lean version and budget. To our knowledge, no existing method optimizes topical impurity, and none optimizes proof quality \textit{during search}. We therefore build baselines atop the strong agent from \citet{nearaiPutnambenchDeepseek}. \textit{+ Instruct} steers the agent during its search: the end of its prompt states the metric and asks for a proof that minimizes it. \textit{+ Refactor} improves the proof after the search, as post-hoc refactoring does: it is handed NearAI's finished baseline proof~(\S~\ref{sec:optimizing-computational-cost}) and asked to improve the proof. Together they cover the two alternatives to optimizing during search:~steering by instruction and repairing afterwards. See Appendix~\ref{appdx:quality-baseline} for details.

\textbf{Reporting.} We use the proofs delivered by NearAI from \S~\ref{sec:optimizing-computational-cost} as a common anchor: we report relative reduction of metric $m$ relative to this baseline proof, averaged over all problems. A run that fails or exceeds its budget is scored at the same metric as the baseline and charged at \$4.5. Finally, because \emph{+ Refactor} starts from the NearAI proof baseline, its cost is the original cost plus the cost to refactor.

\begin{figure}[ht]
      \centering
      \includegraphics[width=\linewidth]{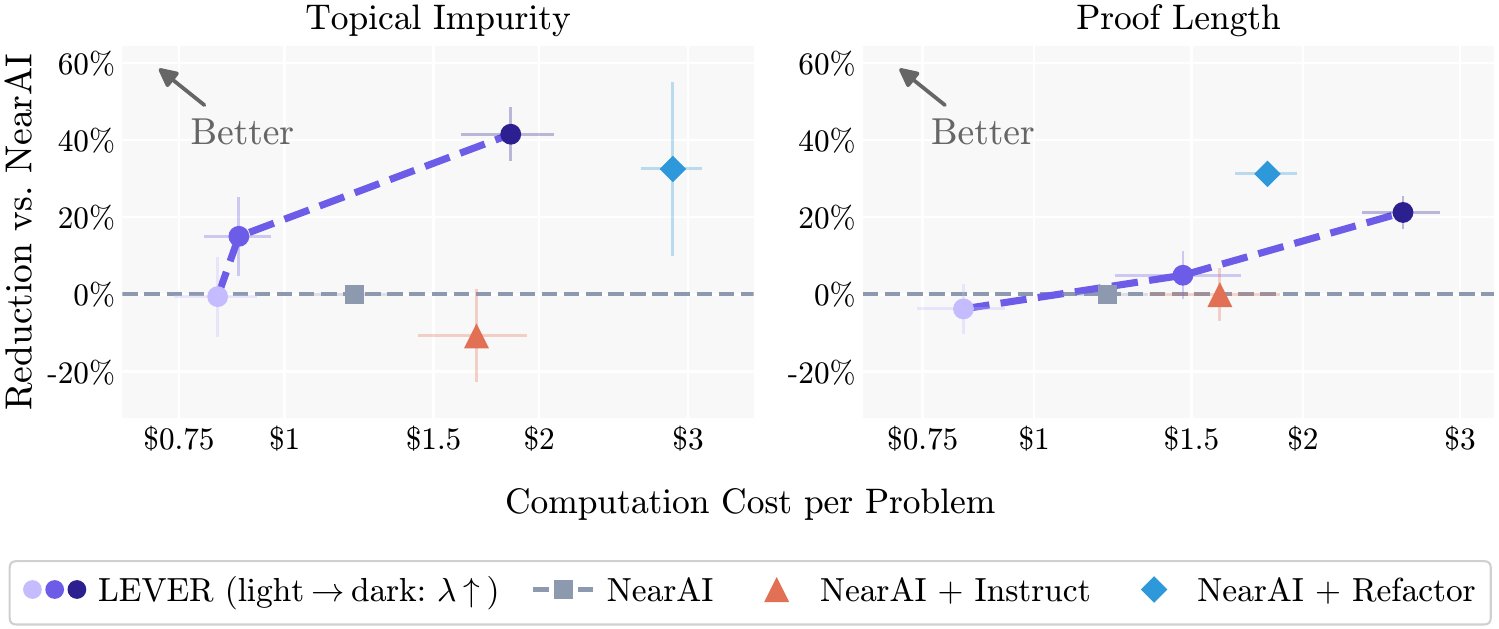}
      \caption{\textbf{Proof quality against computation cost.} Mean reduction of each metric relative to NearAI's proof of the same problem, against mean cost per problem, for topical impurity (left) and proof length
  (right), on the same $20$ problems. \lever's points are $\lambda = 0, 0.01, 0.1$ for impurity and $\lambda = 0, 10^{-4}, 10^{-3}$ for length, light to dark; $\lambda = 0$ is \lever run with its \$3 budget and routing to direct attempts (\S~\ref{sec:optimizing-computational-cost}); the $\lambda > 0$ runs have \$4.5 and no routing. A run that fails or exceeds \$4.5 is scored as NearAI's proof at \$4.5; \textit{+ Refactor} is also charged for NearAI's run. Error bars: $\pm 1$ standard error over
  problems.}
      \label{fig:quality-pareto}
  \end{figure}

\begin{table}[t]
\centering
\begin{tabular}{l rr rr}
\toprule
  & \multicolumn{2}{c}{$m = \texttt{imp}$} & \multicolumn{2}{c}{$m = \texttt{len}$} \\
\cmidrule(lr){2-3}\cmidrule(lr){4-5}
Method & $\Delta$Impurity & $\Delta$Length & $\Delta$Length & $\Delta$Impurity \\
\midrule
\lever ($\lambda{=}0$)   & $+1\%$  & $+4\%$  & $+4\%$  & $+1\%$ \\
  \textsc{Lever}$^{\lambda_1}_m$     & $-15\%$ & $+8\%$  & $-5\%$  & $+3\%$ \\
  \textsc{Lever}$^{\lambda_2}_m$     & $-42\%$ & $+10\%$ & $-21\%$ & $-18\%$ \\
\bottomrule
\end{tabular}
\caption{\textbf{Optimizing one metric, measured on both.} Mean change of topical impurity and of proof length relative to NearAI's proof of the same problem; negative is better. $\lambda_1, \lambda_2$ are
$0.01, 0.1$ when optimizing impurity and $10^{-4}, 10^{-3}$ when optimizing length. A run that fails or exceeds \$4.5 is scored as NearAI's proof.}
\label{tab:quality-pairwise}
\end{table}

\textbf{Length optimization.} We run \textsc{Lever}$_\texttt{len}^\lambda$ for $\lambda \in \{0.0001, 0.001\}$ and plot the results in Figure~\ref{fig:quality-pareto} (right). The proof length objective is well optimized by refactoring: previous works have independently reported trained refactoring models shorten PutnamBench proofs by 57\%~\citep{gu2025proofoptimizertraininglanguagemodels} and competition proofs by over 70\%~\citep{leanrefactor}, in their respective settings. Our \textit{+ Refactor} baseline agrees: it shortens NearAI's proofs by $31\%$ at \$1.83 per problem. At $\lambda = 10^{-3}$, \lever approaches \textit{+ Refactor}: its proofs are $21\%$ shorter, though at a higher
  cost of \$2.59, indicating that searching for short proofs may be more challenging than searching for a correct proof first followed by refactoring.

\textbf{Impurity optimization.} We run \textsc{Lever}$_\texttt{imp}^\lambda$ for $\lambda \in \{0.01, 0.1\}$ with results in Figure~\ref{fig:quality-pareto} (left). Unlike length, refactoring is not built for topical impurity because removing an out-of-subject reference usually means changing the argument rather than tidying it. Our \textit{+ Refactor} baseline reduces impurity by $33\%$ at \$2.88 per problem. At $\lambda = 0.1$, \lever's proofs are $42\%$ purer, at \$1.85, about two-thirds of \textit{+ Refactor}'s cost. \lever is also more reliable: the standard deviation of its per-problem reduction is $32$ points, against $101$ for \textit{+ Refactor}, whose rewrites can backfire badly (on \texttt{putnam\_2007\_a2}, from $29$
  out-of-subject references to $132$). Instructing the agent to look for pure proofs does
  not help either: \textit{+ Instruct}'s proofs are, on the contrary, $11\%$ less pure than the baseline, indicating that the agent loses context of the search objective in its single long session. This indicates that, when a metric depends on the proof strategy itself, optimizing it during search and treating it locally, works best.

\textbf{Trade-offs among metrics.} As shown in Figure~\ref{fig:quality-pareto}, \lever allows tracing a cost-quality trade-off curve: as $\lambda$ rises, proof quality improves and the search costs more. Since $\lambda$ is a conversion rate between metrics, a user can choose how much quality is worth paying for, and move smoothly along this frontier. In contrast, any fixed policy, such as the baselines considered here, admits only single points in this space. We observe that \lever's search also reveals relationships among various quality metrics~(Table~\ref{tab:quality-pairwise}). Shortening a proof makes it purer: as proofs get shorter ($+4\% \rightarrow -21\%$), impurity reduces as well ($+1\% \rightarrow -18\%$). This is likely because as a proof is shortened, impure references are removed
  for free. However, the reverse does not hold: optimizing for impurity alone ($+1\% \rightarrow -42\%$) makes proofs longer ($+4\% \rightarrow +10\%$); this supports our observation that optimizing impurity requires a
  different strategy than what works for length reduction.

\section{Conclusion} \label{sec:conclusion}

We presented \lever, a proof search algorithm that treats correctness as a constraint and makes the objective over correct proofs programmable. By valuing partial proofs over an \andor graph, combining realized costs with predictions for open goals, \lever optimizes that objective during search rather than repairing a proof afterwards. On PutnamBench, it is $34\%$ cheaper than a strong single-conversation agent while solving more problems ($96\%$ against $80\%$). On topical impurity, search-time optimization removes more out-of-subject references than post-hoc refactoring at two-thirds of its cost, and on proof length it approaches refactoring. A single weight trades quality against cost, letting a user choose how much a better proof is worth.

\subsection*{Generative AI Use Statement}

Beyond the LLMs that are the object of our experiments, we used generative AI tools to write code, run and monitor experiments, survey the literature, check ideas, draft proofs, and polish the writing. AI tools had no or minimal involvement in developing the main ideas of the paper, or in interpreting and iterating on experimental results. We have reviewed AI-assisted work to our best ability, and we take responsibility for the final content of this paper, including text, claims and artifacts produced with the aid of generative AI.

\subsection*{Acknowledgments}

We gratefully acknowledge use of the research computing resources of the Empire AI Consortium, Inc, with support from Empire State Development of the State of New York, the Simons Foundation, and the Secunda Family Foundation~\citep{Bloom2025EmpireAI}.

\bibliography{references}

@inproceedings{Bloom2025EmpireAI,

author    = {Stacie Bloom and Joshua C. Brumberg and Ian Fisk and Robert J. Harrison and Robert Hull and Melur Ramasubramanian and Krystyn Van Vliet and Jeannette Wing},

title     = {Empire {AI}: A new model for provisioning {AI} and {HPC} for academic research in the public good},

booktitle = {Practice and Experience in Advanced Research Computing ({PEARC} '25)},

pages     = {4},

year      = {2025},

month     = {July},

address   = {Columbus, OH, USA},

publisher = {ACM},

location  = {New York, NY, USA},

doi       = {10.1145/3708035.3736070},

url       = {https://doi.org/10.1145/3708035.3736070}

}

@article{sahni1974computationally,
      title   = {Computationally Related Problems},
      author  = {Sahni, Sartaj},
      journal = {SIAM Journal on Computing},
      volume  = {3},
      number  = {4},
      pages   = {262--279},
      year    = {1974},
      doi     = {10.1137/0203021}
}

@article{weitzman1979optimal,
author = {Martin L. Weitzman},
title = {Optimal Search for the Best Alternative},
journal = {Econometrica},
volume = {47},
number = {3},
pages = {641-654},
year = {1979}
}

@misc{desabbata2024rational,
      title={Rational Metareasoning for Large Language Models}, 
      author={C. Nicolò De Sabbata and Theodore R. Sumers and Badr AlKhamissi and Antoine Bosselut and Thomas L. Griffiths},
      year={2025},
      eprint={2410.05563},
      archivePrefix={arXiv},
      primaryClass={cs.CL},
      url={https://arxiv.org/abs/2410.05563}, 
}

@article{russell1991principles,
title = {Principles of metareasoning},
journal = {Artificial Intelligence},
volume = {49},
number = {1},
pages = {361-395},
year = {1991},
issn = {0004-3702},
doi = {https://doi.org/10.1016/0004-3702(91)90015-C},
url = {https://www.sciencedirect.com/science/article/pii/000437029190015C},
author = {Stuart Russell and Eric Wefald}
}

@inproceedings{wang2024mathshepherd,
  author = {Wang, Peiyi and Li, Lei and Shao, Zhihong and Xu, Runxin and Dai, Damai and Li, Yifei and Chen, Deli and Wu, Yu and Sui, Zhifang},
  title = {{Math-Shepherd}: Verify and Reinforce {LLM}s Step-by-step without Human Annotations},
  booktitle = {Proceedings of the 62nd Annual Meeting of the Association for Computational Linguistics (Volume 1: Long Papers)},
  year = {2024},
  pages = {9426–9439},
  publisher = {Association for Computational Linguistics},
  doi = {10.18653/v1/2024.acl-long.510},
  url = {http://dx.doi.org/10.18653/v1/2024.acl-long.510}
}

@misc{lightman2023lets,
      title={Let's Verify Step by Step}, 
      author={Hunter Lightman and Vineet Kosaraju and Yura Burda and Harri Edwards and Bowen Baker and Teddy Lee and Jan Leike and John Schulman and Ilya Sutskever and Karl Cobbe},
      year={2023},
      eprint={2305.20050},
      archivePrefix={arXiv},
      primaryClass={cs.LG},
      url={https://arxiv.org/abs/2305.20050}, 
}

@misc{snell2024scaling,
      title={Scaling {LLM} Test-Time Compute Optimally can be More Effective than Scaling Model Parameters}, 
      author={Charlie Snell and Jaehoon Lee and Kelvin Xu and Aviral Kumar},
      year={2024},
      eprint={2408.03314},
      archivePrefix={arXiv},
      primaryClass={cs.LG},
      url={https://arxiv.org/abs/2408.03314}, 
}

@misc{zhou2024language,
      title={Language Agent Tree Search Unifies Reasoning Acting and Planning in Language Models}, 
      author={Andy Zhou and Kai Yan and Michal Shlapentokh-Rothman and Haohan Wang and Yu-Xiong Wang},
      year={2024},
      eprint={2310.04406},
      archivePrefix={arXiv},
      primaryClass={cs.AI},
      url={https://arxiv.org/abs/2310.04406}, 
}

@inproceedings{hao2023reasoning,
  author = {Hao, Shibo and Gu, Yi and Ma, Haodi and Hong, Joshua and Wang, Zhen and Wang, Daisy and Hu, Zhiting},
  title = {Reasoning with Language Model is Planning with World Model},
  booktitle = {Proceedings of the 2023 Conference on Empirical Methods in Natural Language Processing},
  year = {2023},
  pages = {8154–8173},
  publisher = {Association for Computational Linguistics},
  doi = {10.18653/v1/2023.emnlp-main.507},
  url = {http://dx.doi.org/10.18653/v1/2023.emnlp-main.507}
}

@inproceedings{yao2023tree,
  author = {Yao, Shunyu and Yu, Dian and Zhao, Jeffrey and Shafran, Izhak and Griffiths, Tom and Cao, Yuan and Narasimhan, Karthik},
  title = {Tree of Thoughts: Deliberate Problem Solving with Large Language Models},
  booktitle = {Advances in Neural Information Processing Systems 36},
  year = {2023},
  pages = {11809–11822},
  publisher = {Neural Information Processing Systems Foundation, Inc. (NeurIPS)},
  series = {NeurIPS 2023},
  doi = {10.52202/075280-0517},
  url = {http://dx.doi.org/10.52202/075280-0517},
  collection = {NeurIPS 2023}
}

@inproceedings{chen2020retro,
  author = {Binghong Chen and Chengtao Li and Hanjun Dai and Le Song},
  title = {Retro*: Learning Retrosynthetic Planning with Neural Guided {A}* Search},
  booktitle = {Proceedings of the 37th International Conference on Machine Learning},
  year = {2020}
}

@misc{lample2022hypertree,
      title={{HyperTree} Proof Search for Neural Theorem Proving}, 
      author={Guillaume Lample and Marie-Anne Lachaux and Thibaut Lavril and Xavier Martinet and Amaury Hayat and Gabriel Ebner and Aurélien Rodriguez and Timothée Lacroix},
      year={2022},
      eprint={2205.11491},
      archivePrefix={arXiv},
      primaryClass={cs.CL},
      url={https://arxiv.org/abs/2205.11491}, 
}

@article{silver2016mastering,
  author = {Silver, David and Huang, Aja and Maddison, Chris J. and Guez, Arthur and Sifre, Laurent and van den Driessche, George and Schrittwieser, Julian and Antonoglou, Ioannis and Panneershelvam, Veda and Lanctot, Marc and Dieleman, Sander and Grewe, Dominik and Nham, John and Kalchbrenner, Nal and Sutskever, Ilya and Lillicrap, Timothy and Leach, Madeleine and Kavukcuoglu, Koray and Graepel, Thore and Hassabis, Demis},
  title = {Mastering the game of {G}o with deep neural networks and tree search},
  journal = {Nature},
  year = {2016},
  month = {Jan},
  volume = {529},
  number = {7587},
  pages = {484–489},
  publisher = {Springer Science and Business Media LLC},
  doi = {10.1038/nature16961},
  url = {http://dx.doi.org/10.1038/nature16961},
  issn = {1476-4687}
}

@article{silver2018general,
  author = {Silver, David and Hubert, Thomas and Schrittwieser, Julian and Antonoglou, Ioannis and Lai, Matthew and Guez, Arthur and Lanctot, Marc and Sifre, Laurent and Kumaran, Dharshan and Graepel, Thore and Lillicrap, Timothy and Simonyan, Karen and Hassabis, Demis},
  title = {A general reinforcement learning algorithm that masters chess, shogi, and {G}o through self-play},
  journal = {Science},
  year = {2018},
  month = {Dec},
  volume = {362},
  number = {6419},
  pages = {1140–1144},
  publisher = {American Association for the Advancement of Science (AAAS)},
  doi = {10.1126/science.aar6404},
  url = {http://dx.doi.org/10.1126/science.aar6404},
  issn = {1095-9203}
}

@InProceedings{kocsis2006bandit,
author="Kocsis, Levente
and Szepesv{\'a}ri, Csaba",
editor="F{\"u}rnkranz, Johannes
and Scheffer, Tobias
and Spiliopoulou, Myra",
title="Bandit Based {M}onte-{C}arlo Planning",
booktitle="Machine Learning: ECML 2006",
year="2006",
publisher="Springer Berlin Heidelberg",
address="Berlin, Heidelberg",
pages="282--293",
isbn="978-3-540-46056-5"
}

@article{allis1994proof,
author = {Allis, L. Victor and van der Meulen, Maarten and van den Herik, H. Jaap},
title = {Proof-number search},
year = {1994},
issue_date = {March 1994},
publisher = {Elsevier Science Publishers Ltd.},
address = {GBR},
volume = {66},
number = {1},
issn = {0004-3702},
url = {https://doi.org/10.1016/0004-3702(94)90004-3},
doi = {10.1016/0004-3702(94)90004-3},
journal = {Artif. Intell.},
month = mar,
pages = {91–124},
numpages = {34}
}

@article{hansen2001lao,
title = {{LAO}$^*$: A heuristic search algorithm that finds solutions with loops},
journal = {Artificial Intelligence},
volume = {129},
number = {1},
pages = {35-62},
year = {2001},
issn = {0004-3702},
doi = {https://doi.org/10.1016/S0004-3702(01)00106-0},
url = {https://www.sciencedirect.com/science/article/pii/S0004370201001060},
author = {Eric A. Hansen and Shlomo Zilberstein}
}

@inproceedings{martelli1973additive,
  title     = {{Additive AND/OR Graphs}},
  author    = {Martelli, Alberto and Montanari, Ugo},
  booktitle = {International Joint Conference on Artificial Intelligence},
  year      = {1973},
  pages     = {1-11},
  url       = {https://mlanthology.org/ijcai/1973/martelli1973ijcai-additive/}
}

@misc{quarry2026,
      title={Planning to Hammer: Difficulty-Aware Decomposition for Automating {R}ocq Proofs}, 
      author={Ning Zhang and Nongyu Di and Zenan Li and Yuan Yao and Xiaoxing Ma},
      year={2026},
      eprint={2606.17981},
      archivePrefix={arXiv},
      primaryClass={cs.SE},
      url={https://arxiv.org/abs/2606.17981}, 
}

@article{hubert2025alphaproof,
  author = {Hubert, Thomas and Mehta, Rishi and Sartran, Laurent and Horváth, Miklós Z. and Žužić, Goran and Wieser, Eric and Huang, Aja and Schrittwieser, Julian and Schroecker, Yannick and Masoom, Hussain and Bertolli, Ottavia and Zahavy, Tom and Mandhane, Amol and Yung, Jessica and Beloshapka, Iuliya and Ibarz, Borja and Veeriah, Vivek and Yu, Lei and Nash, Oliver and Lezeau, Paul and Mercuri, Salvatore and Sönne, Calle and Mehta, Bhavik and Davies, Alex and Zheng, Daniel and Pedregosa, Fabian and Li, Yin and von Glehn, Ingrid and Rowland, Mark and Albanie, Samuel and Velingker, Ameya and Schmitt, Simon and Lockhart, Edward and Hughes, Edward and Michalewski, Henryk and Sonnerat, Nicolas and Hassabis, Demis and Kohli, Pushmeet and Silver, David},
  title = {Olympiad-level formal mathematical reasoning with reinforcement learning},
  journal = {Nature},
  year = {2025},
  month = {Nov},
  volume = {651},
  number = {8106},
  pages = {607–613},
  publisher = {Springer Science and Business Media LLC},
  doi = {10.1038/s41586-025-09833-y},
  url = {http://dx.doi.org/10.1038/s41586-025-09833-y},
  issn = {1476-4687}
}

@misc{chen2025seedprover,
      title={{Seed-Prover}: Deep and Broad Reasoning for Automated Theorem Proving}, 
      author={Luoxin Chen and Jinming Gu and Liankai Huang and Wenhao Huang and Zhicheng Jiang and Allan Jie and Xiaoran Jin and Xing Jin and Chenggang Li and Kaijing Ma and Cheng Ren and Jiawei Shen and Wenlei Shi and Tong Sun and He Sun and Jiahui Wang and Siran Wang and Zhihong Wang and Chenrui Wei and Shufa Wei and Yonghui Wu and Yuchen Wu and Yihang Xia and Huajian Xin and Fan Yang and Huaiyuan Ying and Hongyi Yuan and Zheng Yuan and Tianyang Zhan and Chi Zhang and Yue Zhang and Ge Zhang and Tianyun Zhao and Jianqiu Zhao and Yichi Zhou and Thomas Hanwen Zhu},
      year={2025},
      eprint={2507.23726},
      archivePrefix={arXiv},
      primaryClass={cs.AI},
      url={https://arxiv.org/abs/2507.23726}, 
}

@misc{ren2025deepseekproverv2,
      title={{DeepSeek-Prover-V2}: Advancing Formal Mathematical Reasoning via Reinforcement Learning for Subgoal Decomposition}, 
      author={Z. Z. Ren and Zhihong Shao and Junxiao Song and Huajian Xin and Haocheng Wang and Wanjia Zhao and Liyue Zhang and Zhe Fu and Qihao Zhu and Dejian Yang and Z. F. Wu and Zhibin Gou and Shirong Ma and Hongxuan Tang and Yuxuan Liu and Wenjun Gao and Daya Guo and Chong Ruan},
      year={2025},
      eprint={2504.21801},
      archivePrefix={arXiv},
      primaryClass={cs.CL},
      url={https://arxiv.org/abs/2504.21801}, 
}

@misc{lin2025goedelproverv2,
      title={{Goedel-Prover-V2}: Scaling Formal Theorem Proving with Scaffolded Data Synthesis and Self-Correction}, 
      author={Yong Lin and Shange Tang and Bohan Lyu and Ziran Yang and Jui-Hui Chung and Haoyu Zhao and Lai Jiang and Yihan Geng and Jiawei Ge and Jingruo Sun and Jiayun Wu and Jiri Gesi and Ximing Lu and David Acuna and Kaiyu Yang and Hongzhou Lin and Yejin Choi and Danqi Chen and Sanjeev Arora and Chi Jin},
      year={2025},
      eprint={2508.03613},
      archivePrefix={arXiv},
      primaryClass={cs.LG},
      url={https://arxiv.org/abs/2508.03613}, 
}

@misc{wang2025kimina,
      title={{Kimina-Prover} Preview: Towards Large Formal Reasoning Models with Reinforcement Learning}, 
      author={Haiming Wang and Mert Unsal and Xiaohan Lin and Mantas Baksys and Junqi Liu and Marco Dos Santos and Flood Sung and Marina Vinyes and Zhenzhe Ying and Zekai Zhu and Jianqiao Lu and Hugues de Saxcé and Bolton Bailey and Chendong Song and Chenjun Xiao and Dehao Zhang and Ebony Zhang and Frederick Pu and Han Zhu and Jiawei Liu and Jonas Bayer and Julien Michel and Longhui Yu and Léo Dreyfus-Schmidt and Lewis Tunstall and Luigi Pagani and Moreira Machado and Pauline Bourigault and Ran Wang and Stanislas Polu and Thibaut Barroyer and Wen-Ding Li and Yazhe Niu and Yann Fleureau and Yangyang Hu and Zhouliang Yu and Zihan Wang and Zhilin Yang and Zhengying Liu and Jia Li},
      year={2025},
      eprint={2504.11354},
      archivePrefix={arXiv},
      primaryClass={cs.AI},
      url={https://arxiv.org/abs/2504.11354}, 
}

@misc{first2023baldur,
      title={Baldur: Whole-Proof Generation and Repair with Large Language Models}, 
      author={Emily First and Markus N. Rabe and Talia Ringer and Yuriy Brun},
      year={2023},
      eprint={2303.04910},
      archivePrefix={arXiv},
      primaryClass={cs.LG},
      url={https://arxiv.org/abs/2303.04910}, 
}

@misc{xin2025bfsprover,
      title={{BFS-Prover}: Scalable Best-First Tree Search for {LLM}-based Automatic Theorem Proving}, 
      author={Ran Xin and Chenguang Xi and Jie Yang and Feng Chen and Hang Wu and Xia Xiao and Yifan Sun and Shen Zheng and Kai Shen},
      year={2025},
      eprint={2502.03438},
      archivePrefix={arXiv},
      primaryClass={cs.AI},
      url={https://arxiv.org/abs/2502.03438}, 
}

@misc{xin2024deepseekproverv15,
      title={{DeepSeek-Prover-V1.5}: Harnessing Proof Assistant Feedback for Reinforcement Learning and {M}onte-{C}arlo Tree Search}, 
      author={Huajian Xin and Z. Z. Ren and Junxiao Song and Zhihong Shao and Wanjia Zhao and Haocheng Wang and Bo Liu and Liyue Zhang and Xuan Lu and Qiushi Du and Wenjun Gao and Qihao Zhu and Dejian Yang and Zhibin Gou and Z. F. Wu and Fuli Luo and Chong Ruan},
      year={2024},
      eprint={2408.08152},
      archivePrefix={arXiv},
      primaryClass={cs.CL},
      url={https://arxiv.org/abs/2408.08152}, 
}

@misc{yang2023leandojo,
      title={{LeanDojo}: Theorem Proving with Retrieval-Augmented Language Models}, 
      author={Kaiyu Yang and Aidan M. Swope and Alex Gu and Rahul Chalamala and Peiyang Song and Shixing Yu and Saad Godil and Ryan Prenger and Anima Anandkumar},
      year={2023},
      eprint={2306.15626},
      archivePrefix={arXiv},
      primaryClass={cs.LG},
      url={https://arxiv.org/abs/2306.15626}, 
}

@misc{polu2020generative,
      title={Generative Language Modeling for Automated Theorem Proving}, 
      author={Stanislas Polu and Ilya Sutskever},
      year={2020},
      eprint={2009.03393},
      archivePrefix={arXiv},
      primaryClass={cs.LG},
      url={https://arxiv.org/abs/2009.03393}, 
}

@article{yao2023react,
  title={{ReAct}: Synergizing Reasoning and Acting in Language Models},
  author={Yao, Shunyu and Zhao, Jeffrey and Yu, Dian and Du, Nan and Shafran, Izhak and Narasimhan, Karthik and Cao, Yuan},
  journal={arXiv preprint arXiv:2210.03629},
  year={2022}
}

@article{leanfinder,
  title={{Lean Finder}: Semantic search for {M}athlib that understands user intents},
  author={Lu, Jialin and Emond, Kye and Yang, Kaiyu and Chaudhuri, Swarat and Sun, Weiran and Chen, Wuyang},
  journal={arXiv preprint arXiv:2510.15940},
  year={2025}
}

@misc{leanarchitect,
      title={{LeanArchitect}: Automating Blueprint Generation for Humans and {AI}}, 
      author={Thomas Zhu and Pietro Monticone and Jeremy Avigad and Sean Welleck},
      year={2026},
      eprint={2601.22554},
      archivePrefix={arXiv},
      primaryClass={cs.LO},
      url={https://arxiv.org/abs/2601.22554}, 
}

@misc{deepseekv4priceupdate,
  author       = {{DeepSeek-AI}},
  title        = {{DeepSeek-V4-Pro GA Release}},
  year         = {2026},
  month        = aug,
  day          = {13},
  howpublished = {\url{https://api-docs.deepseek.com/news/news260813}},
  note         = {Accessed: 2026-09-21}
}

@misc{inoue2026fourcolor,
title={The Four Color Theorem with Linearly Many Reducible Configurations
and Near-Linear Time Coloring},
author={Inoue, Yuta and Kawarabayashi, Ken-ichi and Miyashita, Atsuyuki and
Mohar, Bojan and Thomassen, Carsten and Thorup, Mikkel},
year={2026},
eprint={2603.24880},
}

@article{gonthier2008formal,
title={Formal proof---the four-color theorem},
author={Gonthier, Georges},
journal={Notices of the AMS},
volume={55},
number={11},
pages={1382--1393},
year={2008}
}

@article{tao2007good,
    title={What is good mathematics?},
    author={Tao, Terence},
    journal={Bulletin of the American Mathematical Society},
    volume={44},
    number={4},
    pages={623--634},
    year={2007}
  }

@article{appel1977every,
    title={Every planar map is four colorable. {Part I}: Discharging},
    author={Appel, Kenneth and Haken, Wolfgang},
    journal={Illinois Journal of Mathematics},
    volume={21},
    number={3},
    pages={429--490},
    year={1977}
  }

@misc{deepseekv42026,
      title={{DeepSeek-V4}: Towards Highly Efficient Million-Token Context Intelligence}, 
      author={DeepSeek-AI and Anyi Xu and Bangcai Lin and Bing Xue and Bingxuan Wang and Bingzheng Xu and Bochao Wu and Bowei Zhang and Chaofan Lin and Chen Dong and Chenchen Ling and Chengda Lu and Chenggang Zhao and Chengqi Deng and Chengyu Hou and Chenhao Xu and Chenze Shao and Chong Ruan and Conner Sun and Damai Dai and Daya Guo and Dejian Yang and Deli Chen and Donghao Li and Dongjie Ji and Erhang Li and Fang Wei and Fangyun Lin and Fangzhou Yuan and Feiyu Xia and Fucong Dai and Guangbo Hao and Guanting Chen and Guoai Cao and Guolai Meng and Guowei Li and Han Yu and Han Zhang and Hanwei Xu and Hao Li and Haofen Liang and Haoling Zhang and Haoming Luo and Haoran Wei and Haotian Yuan and Haowei Zhang and Haowen Luo and Haoyu Chen and Haozhe Ji and Hengqing Zhang and Honghui Ding and Hongxuan Tang and Huanqi Cao and Huazuo Gao and Hui Qu and Hui Zeng and J Yang and JQ Zhu and Jia Luo and Jia Song and Jia Yu and Jialiang Huang and Jialu Cai and Jian Liang and Jiangting Zhou and Jiasheng Ye and Jiashi Li and Jiaxin Xu and Jiewen Hu and Jieyu Yang and Jin Chen and Jin Yan and Jingchang Chen and Jingli Zhou and Jingting Xiang and Jingyang Yuan and Jingyuan Cheng and Jingzi Zhou and Jinhua Zhu and Jiping Yu and Joseph Sun and Jun Ran and Junguang Jiang and Junjie Qiu and Junlong Li and Junmin Zheng and Junxiao Song and Kai Dong and Kaige Gao and Kang Guan and Kexing Zhou and Kezhao Huang and Kuai Yu and Lean Wang and Lecong Zhang and Lei Wang and Leyi Xia and Li Zhang and Liang Zhao and Lihua Guo and Lingxiao Luo and Linwang Ma and Linyan Zhu and Litong Wang and Liyu Cai and Liyue Zhang and Longhao Chen and MS Di and MY Xu and Max Mei and Miaojun Wang and Mingchuan Zhang and Minghua Zhang and Minghui Tang and Mingming Li and Mingxu Zhou and Minmin Han and Ning Wang and Panpan Huang and Panpan Wang and Peixin Cong and Peiyi Wang and Peng Zhang and Qiancheng Wang and Qihao Zhu and Qingyang Li and Qinyu Chen and Qiushi Du and Qiwei Jiang and Rui Tian and Ruifan Xu and Ruijie Lu and Ruiling Xu and Ruiqi Ge and Ruisong Zhang and Ruizhe Pan and Runji Wang and Runqian Chen and Runqiu Yin and Runxin Xu and Ruomeng Shen and Ruoyu Zhang and Ruyi Chen and SH Liu and Shanghao Lu and Shangmian Sun and Shangyan Zhou and Shanhuang Chen and Shaofei Cai and Shaoheng Nie and Shaoqing Wu and Shaoyuan Chen and Shengding Hu and Shengyu Liu and Shiqiang Hu and Shirong Ma and Shiyu Wang and Shuiping Yu and Shunfeng Zhou and Shuting Pan and Shuying Yu and Songyang Zhou and Tao Ni and Tao Yun and Tian Jin and Tian Pei and Tian Ye and Tianle Lin and Tianran Ji and Tianyi Cui and Tianyuan Yue and Tingting Yu and Tun Wang and W Zhang and WL Xiao and Wangding Zeng and Wei An and Weilin Zhao and Wen Liu and Wenfeng Liang and Wenjie Pang and Wenjing Luo and Wenjing Yao and Wenjun Gao and Wenkai Yang and Wenlve Huang and Wenqing Hou and Wentao Zhang and Wenting Ma and Xi Gao and Xiang He and Xiangwen Wang and Xianzu Wang and Xiao Bi and Xiaodong Liu and Xiaohan Wang and Xiaokang Chen and Xiaokang Zhang and Xiaotao Nie and Xiaowen Sun and Xiaoxiang Wang and Xin Cheng and Xin Liu and Xin Xie and Xingchao Liu and Xingchen Liu and Xingkai Yu and Xingyou Li and Xinyu Yang and Xinyu Zhang and Xu Chen and Xuanyu Wang and Xuecheng Su and Xueyin Chen and Xuheng Lin and Xuwei Fu and YC Yan and YQ Wang and YW Ma and Yanfeng Luo and Yang Zhang and Yanhong Xu and Yanru Ma and Yanwen Huang and Yao Li and Yao Li and Yao Xu and Yao Zhao and Yaofeng Sun and Yaohui Wang and Yi Qian and Yi Shao and Yi Yu and Yichao Zhang and Yifan Ding and Yifan Shi and Yijia Wu and Yiliang Xiong and Yiling Ma and Ying He and Ying Tang and Ying Zhou and Yingjia Luo and Yinmin Zhong and Yishi Piao and Yisong Wang and Yixiang Zhang and Yixiao Chen and Yixuan Tan and Yixuan Wei and Yiyang Ma and Yiyuan Liu and Yonglun Yang and Yongqiang Guo and Yongtong Wu and Yu Wu and YuKun Li and Yuan Cheng and Yuan Ou and Yuanfan Xu and Yuanhao Li and Yuduan Wang and Yuehan Yang and Yuer Xu and Yuhan Wu and Yuhao Meng and Yuheng Zou and Yukun Zha and Yunfan Xiong and Yupeng Chen and Yuping Lin and Yuqian Cao and Yuqian Wang and Yushun Zhang and Yuting Yan and Yutong Lin and Yuxian Gu and Yuxiang Luo and Yuxiang You and Yuxuan Liu and Yuxuan Zhou and Yuyang Zhou and Yuzhen Huang and ZF Wu and Zehao Wang and Zehua Zhao and Zehui Ren and Zekai Zhang and Zhangli Sha and Zhe Fu and Zhe Ju and Zhean Xu and Zhenda Xie and Zhengyan Zhang and Zheren Gao and Zhewen Hao and Zhibin Gou and Zhicheng Ma and Zhigang Yan and Zhihong Shao and Zhixian Huang and Zhixuan Chen and Zhiyu Wu and Zhizhou Ren and Zhongyu Wu and Zhuoshu Li and Zhuping Zhang and Zian Xu and Zihao Wang and Zihua Qu and Zihui Gu and Zijia Zhu and Zilin Li and Zipeng Zhang and Ziwei Xie and Ziyi Gao and Ziyi Wan and Zizheng Pan and Zongqing Yao},
      year={2026},
      eprint={2606.19348},
      archivePrefix={arXiv},
      primaryClass={cs.CL},
      url={https://arxiv.org/abs/2606.19348}, 
}

@inproceedings{The_mathlib_Community_2020, series={POPL ’20},
   title={The {L}ean mathematical library},
   url={http://dx.doi.org/10.1145/3372885.3373824},
   DOI={10.1145/3372885.3373824},
   booktitle={Proceedings of the 9th ACM SIGPLAN International Conference on Certified Programs and Proofs},
   publisher={ACM},
   author={{Community Mathlib}},
   year={2020},
   month=Jan, pages={367–381},
   collection={POPL ’20} }

@misc{gu2025proofoptimizertraininglanguagemodels,
      title={{ProofOptimizer}: Training Language Models to Simplify Proofs without Human Demonstrations}, 
      author={Alex Gu and Bartosz Piotrowski and Fabian Gloeckle and Kaiyu Yang and Aram H. Markosyan},
      year={2025},
      eprint={2510.15700},
      archivePrefix={arXiv},
      primaryClass={cs.LG},
      url={https://arxiv.org/abs/2510.15700}, 
}

@article{thiele2003hilbert,
  title={Hilbert's twenty-fourth problem},
  author={Thiele, R{\"u}diger},
  journal={The American Mathematical Monthly},
  volume={110},
  number={1},
  pages={1--24},
  year={2003},
  publisher={Taylor \& Francis}
}

@misc{nearaiPutnambenchDeepseek,
  author       = {NearAI},
  title        = {putnambench-deepseek: An autonomous {DeepSeek}-driven agent
                  that solved {PutnamBench} in {Lean}~4},
  year         = {2026},
  publisher    = {GitHub},
  journal      = {GitHub repository},
  howpublished = {\url{https://github.com/nearai/putnambench-deepseek}},
  note         = {Accessed 14 September 2026}
}

@article{steiner1978,
  title={Mathematical explanation},
  author={Steiner, Mark},
  journal={Philosophical Studies: An International Journal for Philosophy in the Analytic Tradition},
  volume={34},
  number={2},
  pages={135--151},
  year={1978},
  publisher={JSTOR}
}

@article{detlefsen2011purity,
  title={Purity of methods},
  author={Detlefsen, Michael and Arana, Andrew},
  journal={Philosophers' Imprint},
  volume={11},
  year={2011}
}

@misc{improver,
      title={{ImProver}: Agent-Based Automated Proof Optimization}, 
      author={Riyaz Ahuja and Jeremy Avigad and Prasad Tetali and Sean Welleck},
      year={2024},
      eprint={2410.04753},
      archivePrefix={arXiv},
      primaryClass={cs.AI},
      url={https://arxiv.org/abs/2410.04753}, 
}

@misc{fu2026proofrefactorrefactoringgeneratedformal,
      title={{Proof-Refactor}: Refactoring Generated Formal Proofs into Modular Artifacts}, 
      author={Yiming Fu and Peixuan Liu and Zichen Wang and Kun Yuan},
      year={2026},
      eprint={2606.03743},
      archivePrefix={arXiv},
      primaryClass={cs.AI},
      url={https://arxiv.org/abs/2606.03743}, 
}

@misc{leanrefactor,
      title={{Lean Refactor}: Multi-Objective Controllable Proof Optimization via Agentic Strategy Search}, 
      author={Jialin Lu and Soonho Kong and Rodrigo Stehling and Kaiyu Yang and Zhangyang Wang and Weiran Sun and Wuyang Chen},
      year={2026},
      eprint={2605.20244},
      archivePrefix={arXiv},
      primaryClass={cs.LO},
      url={https://arxiv.org/abs/2605.20244}, 
}

@article{kung2026leap,
  title={{LEAP}: Supercharging {LLM}s for Formal Mathematics with Agentic Frameworks},
  author={Kung, Po-Nien and Song, Linfeng and Hwang, Dawsen and Yoon, Jinsung and Li, Chun-Liang and Severini, Simone and Ol{\v{s}}{\'a}k, Mirek and Lockhart, Edward and Le, Quoc V and Gokturk, Burak and others},
  journal={arXiv preprint arXiv:2606.03303},
  year={2026}
}

@misc{tsoukalas2026alphaproofnexus,
      title={Advancing Mathematics Research with {AI}-Driven Formal Proof Search}, 
      author={George Tsoukalas and Anton Kovsharov and Sergey Shirobokov and Anja Surina and Moritz Firsching and Gergely Bérczi and Francisco J. R. Ruiz and Arun Suggala and Adam Zsolt Wagner and Eric Wieser and Lei Yu and Aja Huang and Miklós Z. Horváth and Andrew Ferraiuolo and Henryk Michalewski and Edward Lockhart and Codrut Grosu and Thomas Hubert and Matej Balog and Pushmeet Kohli and Swarat Chaudhuri},
      year={2026},
      eprint={2605.22763},
      archivePrefix={arXiv},
      primaryClass={cs.AI},
      url={https://arxiv.org/abs/2605.22763}, 
}

@inproceedings{
dong2024formal,
title={Formal Theorem Proving by Rewarding {LLM}s to Decompose Proofs Hierarchically},
author={Kefan Dong and Arvind V. Mahankali and Tengyu Ma},
booktitle={The 4th Workshop on Mathematical Reasoning and AI at NeurIPS'24},
year={2024},
url={https://openreview.net/forum?id=D83tiHiNfF}
}

@inproceedings{
wang2024legoprover,
title={{LEGO-Prover}: Neural Theorem Proving with Growing Libraries},
author={Haiming Wang and Huajian Xin and Chuanyang Zheng and Zhengying Liu and Qingxing Cao and Yinya Huang and Jing Xiong and Han Shi and Enze Xie and Jian Yin and Zhenguo Li and Xiaodan Liang},
booktitle={The Twelfth International Conference on Learning Representations},
year={2024},
url={https://openreview.net/forum?id=3f5PALef5B}
}

@inproceedings{
jiang2023draft,
title={Draft, Sketch, and Prove: Guiding Formal Theorem Provers with Informal Proofs},
author={Albert Qiaochu Jiang and Sean Welleck and Jin Peng Zhou and Timothee Lacroix and Jiacheng Liu and Wenda Li and Mateja Jamnik and Guillaume Lample and Yuhuai Wu},
booktitle={The Eleventh International Conference on Learning Representations },
year={2023},
url={https://openreview.net/forum?id=SMa9EAovKMC}
}

@inproceedings{
tsoukalas2024putnambench,
title={{PutnamBench}: Evaluating Neural Theorem-Provers on the {P}utnam Mathematical Competition},
author={George Tsoukalas and Jasper Lee and John Jennings and Jimmy Xin and Michelle Ding and Michael Jennings and Amitayush Thakur and Swarat Chaudhuri},
booktitle={The Thirty-eight Conference on Neural Information Processing Systems Datasets and Benchmarks Track},
year={2024},
url={https://openreview.net/forum?id=ChKCF75Ocd}
}

@article{martinot2024purity,
	author = {Robin Martinot},
	doi = {10.1017/s1755020323000333},
	journal = {Review of Symbolic Logic},
	number = {2},
	pages = {395--434},
	title = {Ontological Purity for Formal Proofs},
	volume = {17},
	year = {2024}
}

@misc{rögnvaldsson2026optimizingcostqualitytradeoffagentic,
      title={Optimizing the Cost-Quality Tradeoff of Agentic Theorem Provers in {L}ean}, 
      author={Kári Rögnvaldsson and Chenhao Sun and Jasper Dekoninck and Martin Vechev},
      year={2026},
      eprint={2606.04883},
      archivePrefix={arXiv},
      primaryClass={cs.CL},
      url={https://arxiv.org/abs/2606.04883}, 
}

@book{nilsson2014principles,
  title={Principles of artificial intelligence},
  author={Nilsson, Nils J},
  year={2014},
  publisher={Morgan Kaufmann}
}

@misc{MistralAI_2026,
  author = {MistralAI},
  title = {LeanStral},
  howpublished = {\url{https://github.com/mistralai/LeanstralSafeVerify/blob/main/LeanstralReport.pdf}},
  year = {2026},
  month = {Jul}
}

@misc{achim2025aristotleimolevelautomatedtheorem,
  title         = {Aristotle: {IMO}-level Automated Theorem Proving},
  author        = {Tudor Achim and Alex Best and Alberto Bietti and Kevin Der
                   and Mathïs Fédérico and Sergei Gukov
                   and Daniel Halpern-Leistner and Kirsten Henningsgard
                   and Yury Kudryashov and Alexander Meiburg
                   and Martin Michelsen and Riley Patterson
                   and Eric Rodriguez and Laura Scharff
                   and Vikram Shanker and Vladmir Sicca
                   and Hari Sowrirajan and Aidan Swope and Matyas Tamas
                   and Vlad Tenev and Jonathan Thomm
                   and Harold Williams and Lawrence Wu},
  year          = {2025},
  eprint        = {2510.01346},
  archivePrefix = {arXiv},
  primaryClass  = {cs.AI},
  url           = {https://arxiv.org/abs/2510.01346}
}

@misc{hilbert,
      title={Hilbert: Recursively Building Formal Proofs with Informal Reasoning}, 
      author={Sumanth Varambally and Thomas Voice and Yanchao Sun and Zhifeng Chen and Rose Yu and Ke Ye},
      year={2026},
      eprint={2509.22819},
      archivePrefix={arXiv},
      primaryClass={cs.AI},
      url={https://arxiv.org/abs/2509.22819}, 
}

@misc{chung2026goedelarchitectstreamliningformaltheorem,
      title={{Goedel-Architect}: Streamlining Formal Theorem Proving with Blueprint Generation and Refinement}, 
      author={Jui-Hui Chung and Ziyang Cai and Zihao Li and Qishuo Yin and Rohit Agarwal and Simon Park and Rodrigo Porto and Narutatsu Ri and Ziran Yang and Shange Tang and Xingyu Dang and Hongzhou Lin and Mengdi Wang and Danqi Chen and Chi Jin and Liam H Fowl and Sanjeev Arora},
      year={2026},
      eprint={2606.06468},
      archivePrefix={arXiv},
      primaryClass={cs.AI},
      url={https://arxiv.org/abs/2606.06468}, 
}

@misc{santos2025kiminaleanserverhighperformance,
      title={{Kimina Lean Server}: A High-Performance {L}ean Server for Large-Scale Verification}, 
      author={Marco Dos Santos and Hugues de Saxcé and Haiming Wang and Ran Wang and Mantas Baksys and Mert Unsal and Junqi Liu and Zhengying Liu and Jia Li},
      year={2025},
      eprint={2504.21230},
      archivePrefix={arXiv},
      primaryClass={cs.LO},
      url={https://arxiv.org/abs/2504.21230}, 
}

@misc{guez2018learningsearchmctsnets,
      title={Learning to Search with {MCTSnets}}, 
      author={Arthur Guez and Théophane Weber and Ioannis Antonoglou and Karen Simonyan and Oriol Vinyals and Daan Wierstra and Rémi Munos and David Silver},
      year={2018},
      eprint={1802.04697},
      archivePrefix={arXiv},
      primaryClass={cs.AI},
      url={https://arxiv.org/abs/1802.04697}, 
}
\bibliographystyle{iclr2027_conference}

\newpage
\appendix
\section{\lever: System Design} \label{appdx:system-design}

\textbf{Planner.} Before the search, one planner conversation per problem writes an informal proof. It has \texttt{search\_mathlib} and \texttt{python}, runs at high reasoning effort, and is capped at \$0.20. The plan is written once per problem and reused by every \lever run on that problem. Every expansion carries it in its system prompt, and is told that the plan is a guide, which the task in hand supersedes where the two disagree.

\textbf{Tools.} An expansion edits the goal's Lean file in a buffer (\texttt{view}, \texttt{replace}, \texttt{replace\_all}, \texttt{delete\_lines}), and any edit can compile the file in the same step. It can also check a Lean snippet without touching the buffer (\texttt{lean\_check}), search Mathlib (\texttt{search\_mathlib}), run Python (\texttt{python}), compile the file against every submission rule (\texttt{lean\_compile}), and submit the last file that compiled (\texttt{submit\_proof}). A subgoal's conversation may instead report its goal false (\texttt{disprove}).

\textbf{Decomposing a goal.} The agent writes the proof of the target in the file it is given, and may add two kinds of helper lemmas above it. \emph{Closed} helpers are small steps proved in place. \emph{Open} helpers are large steps left as \texttt{sorry\_using [\dots]}, where the list names the declarations in the file that the helper's own proof will use; each becomes a subgoal, proved in a later conversation. Every declaration it adds carries a LeanArchitect blueprint attribute, an informal statement and proof in words, so a subgoal's conversation knows what its lemma is for. A file is accepted only if its header, the declarations it was handed and the target's signature are unchanged; the target is its last declaration, with no \texttt{sorry} in its body; every helper lemma is either proved or open with exactly \texttt{sorry\_using [\dots]} as its body; and it uses none of \texttt{native\_decide}, \texttt{axiom}, \texttt{admit}, or search tactics ending in \texttt{?}.

\textbf{Shared subgoals.} A subgoal's identity is its Lean file with the blueprint annotations stripped: the lemma, together with the open helpers it depends on through \texttt{sorry\_using}. Two decompositions that introduce the same lemma, or a subgoal whose proof uses a sibling it depends on, therefore reach the same OR node, which is proved once. A proof should then pay for a shared subgoal once, not once per use as the recursion of \S\ref{sec:background} does, and under that accounting finding the cheapest solution subgraph is NP-hard~\citep{sahni1974computationally}.
 
\lever instead charges each shared subgoal to one goal, the one whose decomposition first introduced it (or the next, if that decomposition dies). Every alternative decomposition of that goal pays for the subgoal, not only the one that introduced it, so the goal compares its alternatives on what each would cost; decompositions of other goals use the subgoal free, since it is proved once for all of them. Ownership depends only on the order in which nodes were created, not on which alternatives are chosen, so values cannot oscillate; the cost is that a shared subgoal's price drops out of the root's value when its owner chooses an alternative that does not use it.

\textbf{Proving directly.} The direct arm uses the same plan, context and tools, but may leave no helper open.

\textbf{Context inheritance.} A subgoal's conversation starts from its parent's: from the full transcript if the parent's conversation is under $20$K tokens, and otherwise from the parent's compacted output. Any conversation that reaches $128$K tokens is likewise replaced by its compacted output.

\textbf{Redraws.} When a goal is expanded again, its new conversation is told what the earlier attempts found: which of their helpers were proved, which were reported false and why, which were never proved, and what the value function priced the rest at.

\textbf{Disproofs and failures.} A disproof is a claim about the goal rather than the attempt: the goal is closed as false, its value becomes $\infty$, and later decompositions of its parent are told the statement is false and why. An attempt that ends without an accepted file is a failure. Its spend is charged, and its goal's estimate rises by one $\sigma$ (\S\ref{sec:adaptive-proof-search-with-lever}).

\textbf{Parallelism and budget.} Each iteration expands up to $25$ selected goals concurrently, one attempt each. The spend of every LLM call, whether expansion, value function or planner, is charged to the problem's budget as it is made, and running conversations read that spend between calls, so the search stops once the budget is reached rather than after the attempts in flight return.

\textbf{Prompts.} Listings~\ref{lst:plan-prompt}--\ref{lst:prove-prompt} give, verbatim, the planner's prompt, the system prompt of every expansion, and the prompt of a decomposition. \texttt{<statement>} stands for the goal's Lean file, \texttt{<target>} for the declaration to prove, and \texttt{<plan>} for the planner's output. A subgoal's conversation, which continues from its parent's, opens instead with ``The buffer now holds \texttt{<target>}, a lemma you left open above, as its own file, shown below. Your task is to write the proof of \texttt{<target>} in it.'' The direct arm's prompt has the same structure, but requires every new helper to be proved.

\lstdefinestyle{prompt}{
  basicstyle=\fontsize{7}{8.4}\fontencoding{T1}\fontfamily{zi4}\selectfont,
  columns=fullflexible, keepspaces=true, upquote=true,
  breaklines=true, breakindent=0pt,
  frame=single, framerule=0.4pt, xleftmargin=4pt, xrightmargin=4pt,
  captionpos=b,
  literate={—}{{\textemdash}}1,
}

\begin{lstlisting}[style=prompt, caption={The planner's prompt.}, label={lst:plan-prompt}]
Your task is to write a natural language proof of the goal `<target>` found in
the Lean 4 file below.

## File

```lean4
<statement>
```

Your final submission should be the complete mathematical argument that proves
`<target>`, written in natural language prose.

## Instructions

- **Pick a strategy that is easy to formalize**: A later stage will write
  your plan out in Lean 4 against Mathlib, and that is what your plan is judged
  on. So the approach to submit is the one that can be implemented using
  Mathlib machinery that already exists. Every approach may need results
  Mathlib lacks: for each candidate, list what is missing, and pick the one
  whose missing pieces are smallest and most elementary.

- **Survey Mathlib before you commit to an approach**: Use `search_mathlib` to
  investigate what exists in Mathlib, and explicitly name the Mathlib declarations
  that your argument depends on.

- **No Lean proofs, but do name Lean results**: Do NOT write proof tactics or
  segments of a Lean proof. However, you should name the Mathlib declarations
  your steps rest on. Every name you give must be one a search actually
  returned.

- **Verify each step**: Make sure that each step in your argument is logically
  sound and follows from the previous steps. The full proof that you submit should
  be complete and correct, and should not rely on any unproven claims or assumptions.
  Each intermediate claim must be true as a Lean statement using Mathlib, not just
  informally.

- **A missing theory is a reason to change route**: If a step needs a general
  theorem Mathlib does not have, do NOT plan to build that theorem. Look for an
  argument that uses the specific hypotheses you have to prove only the special
  case you need. Build a missing result only when no route avoids it, and then
  build the SMALLEST version your argument uses.

## Tools

You have access to the following tools:

* `search_mathlib`: Use this tool to search for Mathlib declarations that may be
  useful in your proof. Submit queries that describe a Mathlib proposition or
  make a guess at the declaration name (e.g., "the product of two continuous
  functions is continuous" or "Continuous.mul"). Do not describe the problem
  you are solving itself; the index does not hold the answer to the problem.

* `python`: Runs a Python 3 snippet and returns what it printed. The standard
  library, numpy, sympy and scipy are available. Use it to check a claim on
  small cases, or confirm a value before you build the argument on it.
  Use this tool sparingly.

* `submit_plan`: Submits your plan and finishes the task.


## Output Format

Your output should contain one section as follows:

```
# Plan

A step-by-step outline of the argument that proves the statement, each step
justified. Name the Mathlib results the steps rest on. When a step has no
Mathlib result behind it, say so there, and say what you would build it from.

```

## Recommended Workflow

1. Work out how the statement is proved mathematically, and note the two or
three approaches that could work. Do not commit to one yet.
2. Investigate using `search_mathlib` which approach would be the easiest to
formalize in Lean.
3. Work the chosen argument out in full, with each step justified.
4. Verify that each step is sound, then submit it with `submit_plan`.
\end{lstlisting}

\begin{lstlisting}[style=prompt, caption={The system prompt of every expansion.}, label={lst:system-prompt}]
Your goal is to help formalize the solution to the following problem in Lean 4:

```lean4
<statement>
```

Here is the originally constructed plan to solve this problem:

<plan>

Use the above plan to guide your responses for the task shown below.

NOTES:
- The task may have deviated from the plan; in this case the task description
supersedes the plan.
- If something in the plan is false or unclear, you should work out a fix, ignore
the plan and complete the task using your own reasoning. The plan is only a guide,
not a rule.

Your response to the task below should help formalize the solution to the above
problem as easily as possible.
\end{lstlisting}

\begin{lstlisting}[style=prompt, caption={The prompt of a decomposition, as the first expansion of a goal.}, label={lst:prove-prompt}]
Your task is to write the proof of `<target>` in the Lean 4 file below.

The proof may use helper declarations above `<target>`, of two kinds:

- Closed helpers: new theorems/lemmas that you prove here. These represent small
pieces of the argument that are used in the main proof body of `<target>`. You should
prove these here.

- Open helpers: new theorems/lemmas that you leave open using the `sorry_using` tactic.
These represent large pieces of the argument that are used in the main proof body of
`<target>`. These are proved in a later session, so they are expensive. The hardest
open helper you introduce should be much easier than `<target>`. If it is not, an AI
grader will reject your submission. You should be able to judge the quality of your
decomposition by how much work you do in the current file.

## File

```lean4
<statement>
```

## What you submit

The file as it stands when `lean_compile` last accepted it. All of these should
hold, else the file is rejected:

- The header, all declarations above `<target>` and `<target>`'s signature are left
exactly as given (including the blueprint attributes). You may insert new
declarations above `<target>`; do not edit or delete the ones already there.

- `<target>` is the last declaration in the file. Its body contains no `sorry`
or `sorry_using`. If you use `sorry` or `sorry_using` in the main proof body, it
will be rejected but return the open goal(s) as feedback.

- All helper declarations you add and `<target>` carry a blueprint attribute:
```lean4
@[blueprint
  (statement := /-- what it says, in words -/)
  (proof := /-- the mathematical argument, worked out in detail -/)]
```

- Each `lemma` and `theorem` helper is either proved in full (closed), or left open
with exactly `:= by sorry_using [...]` as its whole body, where `[...]` lists other
declarations in the file that the helper will use in its own proof (even
transitively). Do not include Mathlib names or tactics. `[]` if none. When in doubt
whether to add a declaration from this file in the list of dependencies, include it,
aiming for recall.

- All other helpers (like `def` or `abbrev`) should have a complete body and also
carry the blueprint attribute.

- These tactics are not allowed anywhere in the file: `native_decide`, `axiom`,
`admit`, or search tactics ending in `?`.

- Names may collide with a global namespace you cannot see. Rename if told to.


## How to proceed:

- Work out a mathematical argument that can be easily formalized in lean. Ensure
each step in your argument is sound and can be proven in Lean. Consider different
formalization strategies. Your goal is to pick one that is the simplest to formalize
in Lean.

- Search before you commit to an approach. Investigate the Mathlib machinery that
you can reuse instead of doing it yourself. If you are in doubt about a
declaration, use the `search_mathlib` tool to look it up.

- Write the complete proof in the main body, introducing open/closed helpers as
necessary. You should prefer closed helpers and reserve open helpers for only the
hard parts of the proof that are still much easier than `<target>`. If `<target>` can
be proved in this session, prove it HERE.

- Edit the file, compile it, and fix what the compiler reports.

- Never state that the task is too complex or too long to finish. Solve the task
in an iterative manner guided by the compiler feedback.

- When to stop. Use `submit_proof` to finish. It submits the file your last
`lean_compile` ACCEPTED. So if your final revision failed, fix the errors and
compile it successfully before finishing.


## Tools

The file above is open in an editing buffer. You always change it in place:

- `view`/`replace`/`replace_all`/`delete_lines` - use these for interfacing
  with the buffer. Every edit takes `compile: true`, which compiles the file
  right after the edit, exactly as `lean_compile` would, and reports the result
  in the same reply. Set it on the edit that finishes a revision.

- `lean_check` — compiles Lean 4 code you pass to it. This is to be used as a probe
for quick checks of Lean code. Do NOT submit your proof to this tool; use
`lean_compile` for that. Use `lean_check` sparingly. This tool does NOT write to
the buffer.

- `python` — runs a Python 3 snippet and returns what it printed. The standard
library only. Use it to check a claim on small cases, or confirm a value before
you formalise it. Use this tool sparingly.

- `search_mathlib` — describe a proposition, or guess a name ("the product of two
continuous functions is continuous", "Continuous.mul"). You should look for known
mathematical facts that are likely to be present in Mathlib.

- `lean_compile` — checks the file as it now stands against every requirement
above. Reports PASSED, or the failures. Proofs with `sorry` are considered failures but
reveal the goal state where `sorry` occurs, which may be useful for the next attempt.
Run this on your own RARELY. Prefer running compilation after edit as described above.

- `submit_proof` — submits the file your last `lean_compile` accepted, and ends
the task.

IMPORTANT: every response costs a turn, and a turn re-sends the whole
conversation, so put as many tool calls as you can into one response. Calls run in
the order you list them: a later call sees an earlier one's effect (an edit with
`compile: true`, then a `view` of the result, in ONE response) but not its result.
Batch probes together
(`lean_check`, `search_mathlib`, `python`, `view`); make a call on its own only when
its arguments depend on a result you have not seen yet.

Batching tools together is BEST practice and you should practice it as much as possible.
\end{lstlisting}
\section{Topical Impurity} \label{appdx:topical-impurity}

\textbf{Implementation.} Listing~\ref{lst:impurity} computes a proof's impurity. Its input \texttt{uses} comes from \texttt{\#lever\_uses}, a Lean command that elaborates the proof file and reports, for each declaration in order, its kind, whether it rests on \texttt{sorry}, the constants in its statement (\texttt{type}), the library constants in explicit argument positions of its proof (\texttt{mathlib\_refs}), and the file's own declarations its proof references (\texttt{own\_refs}). The last declaration is the theorem being proved. \texttt{TABLE} maps every Mathlib name to its directory, three levels deep, and whether it is a theorem.

\lstdefinestyle{pyimp}{
  language=Python,
  basicstyle=\fontsize{7.5}{9}\fontencoding{T1}\fontfamily{zi4}\selectfont,
  keywordstyle=\bfseries,
  commentstyle=\color{muted},
  stringstyle=\color{black},
  columns=fullflexible, keepspaces=true, upquote=true,
  frame=single, framerule=0.4pt, xleftmargin=4pt, xrightmargin=4pt,
  showstringspaces=false,
}
\begin{lstlisting}[style=pyimp, caption={Topical impurity of a compiled proof, condensed from our implementation (logging and error handling removed).}, label={lst:impurity}, captionpos=b]
ROOT = "<root>"   # the concept of a generic lemma, which belongs to no subject
LOGIC = {"Eq", "Ne", "HEq", "And", "Or", "Iff", "Not", "Exists", "True", "False",
         "Classical", "congrArg", "congrFun", "cast", "id"}
INFRASTRUCTURE = {"Init", "Batteries", "Mathlib/Data/Nat", "Mathlib/Logic/Basic",
                  "Mathlib/Order/Basic", "Mathlib/Order/Defs",
                  "Mathlib/Algebra/BigOperators", "Mathlib/Algebra/Group",
                  "Mathlib/Algebra/GroupWithZero", "Mathlib/Algebra/NeZero",
                  "Mathlib/Algebra/Order", "Mathlib/Algebra/Ring"}

def concept(name):    # the namespace; a bare definition is its own concept
    head, _, rest = name.partition(".")
    if rest:
        return ROOT if head in ("Left", "Right") else head
    return ROOT if TABLE[name].is_theorem else name

def machinery(name):  # tactics, meta code, logical connectives, instances
    where = TABLE[name].location
    return ((name.startswith("inst") and name[4:5].isupper())
            or where.startswith(("Mathlib/Tactic", "Mathlib/Util", "Mathlib/Lean"))
            or where.split("/")[0] in ("Lean", "Aesop", "Qq", "Plausible")
            or concept(name) in LOGIC)

def subject(uses):    # what the statement is about, on both readings
    stated = list(uses.values())[-1]["type"]
    named = [n for n in stated if n not in uses and n in TABLE and not machinery(n)]
    return ({concept(n) for n in named} | {ROOT},
            {TABLE[n].location for n in named})

def through(uses, helpers, seen):  # library names in the helpers' statements
    out = []
    for h in helpers:
        if h in seen:
            continue
        seen.add(h)
        stated = uses[h]["type"]
        out += [n for n in stated if n not in uses]
        out += through(uses, [n for n in stated if n in uses], seen)
    return out

def references(uses):  # once per proved theorem that reaches for a name
    out = []
    for decl in uses.values():
        if decl["kind"] != "theorem" or decl["sorry"]:
            continue
        out += decl["mathlib_refs"] + through(uses, decl["own_refs"], set())
    return out

def impurity(uses):
    concepts, locations = subject(uses)
    count = 0
    for n in references(uses):
        if n not in TABLE or not TABLE[n].is_theorem or machinery(n):
            continue
        where = TABLE[n].location
        if concept(n) == ROOT:
            pure = where in INFRASTRUCTURE or where in locations
        else:
            pure = concept(n) in concepts or where in locations
        count += not pure
    return count
\end{lstlisting}

\textbf{Example.} \texttt{putnam\_1964\_b4} asks how many regions $n$ great circles cut a sphere into. NearAI's proof cites $44$ impure theorems, from topology, convexity and group actions (\eg, \texttt{isOpen\_lt}, \texttt{convex\_iff\_forall\_pos}). \lever's proof at $\lambda = 0.1$ cites $8$, mostly from the linear algebra of the circles' planes (\eg, \texttt{Submodule.finrank\_sup\_add\_finrank\_inf\_eq}).

\textbf{Can the metric be gamed?} A proof could lower its count by re-proving an out-of-subject library lemma as a local helper: the helper's statement names only definitions, which are not charged, and its proof may cite nothing impure. We checked for this on the $20$ problems of \S\ref{sec:proof-quality-optimization}, counting the helpers whose statements name something outside the subject. At $\lambda = 0.1$, \lever writes more helpers than at $\lambda = 0$ ($36.1$ against $25.9$ per proof), but only slightly more off-subject ones ($11.2$ against $10.1$; NearAI: $14.1$). Charging every off-subject helper as one more impure reference lowers \lever's mean impurity reduction at $\lambda = 0.1$ by under one point.

\textbf{Limitations.} Topical impurity is a proxy for purity of method. It takes Mathlib's organization as a stand-in for mathematical subject, and namespace and directory boundaries do not always match a mathematician's; the infrastructure set is a judgment, measured on one pilot corpus. It counts references rather than weighing them, so an essential detour through another field counts the same as a routine lemma from it. And, as above, it charges a local re-proof of an out-of-subject lemma only through the library theorems that re-proof cites.
\section{Experimental Details} \label{appdx:experiments}
\subsection{Problem Selection} \label{appdx:data}

We rank PutnamBench with the published run of \citet{nearaiPutnambenchDeepseek}, which solves all $672$ problems and records every attempt's tokens and wall-clock time. Each problem is ranked by the token cost of its cheapest solving attempt, most expensive first, priced at a fixed rate table of \$0.15, \$0.003 and \$0.60 per million input, cached input and output tokens (prices from \citet{nearaiPutnambenchDeepseek} originally; these rates are no longer available). Most problems are easy for this agent: below rank $150$, the median problem is solved in $0.3$ hours (Table~\ref{tab:selection}). The hardest are impractical for controlled comparisons that run every method on every problem: in the top $50$, $20$ problems needed more than one attempt and one took $5.5$ days. We therefore take ranks $51$--$150$, hard for the agent but solved in a median of $1.5$ hours, and split them at random into $20$ calibration and $80$ evaluation problems.

\begin{table}[h]
    \centering
    \begin{tabular}{l r c}
        \toprule
        Ranks & Problems & Wall-clock, all attempts (h) \\
        \midrule
        1--50    & 50  & 9.6 (max 132) \\
        51--150  & 100 & 1.5 (max 33)  \\
        151--672 & 520 & 0.3 (max 9)   \\
        \bottomrule
    \end{tabular}
    \caption{\textbf{PutnamBench by the cost of the published NearAI run.} Wall-clock: median and maximum over problems of the time summed over all attempts.}
    \label{tab:selection}
\end{table}

\subsection{Value Functions} \label{appdx:value-functions}

Table~\ref{tab:value-functions} lists the constants of \S\ref{sec:adaptive-proof-search-with-lever} for each metric.

\begin{table}[h]
  \centering
  \begin{tabular}{l l l r r r r}
  \toprule
  Metric & Prediction & Scale & $\sigma$ & $\rho$ & $n$ & $\delta$ \\
  \midrule
  \texttt{comp} & LLM, 7 bins & $\log_2$, \$0.016--\$0.512 & 0.615 & 1.08 & 0.33 & 1.85 \\
  \texttt{len}  & LLM, 7 bins & $\log_2$, 512--16{,}384 tokens & 0.408 & 0.72 & 0.32 & 1.50 \\
  \texttt{imp}  & constant ($=4$) & linear & 6.79 & 31.2 & 0.05 & 1.72 \\
  \bottomrule
  \end{tabular}
  \caption{\textbf{Value functions.} $\rho$: prediction error on held-out calibration goals (log scale for \texttt{comp} and \texttt{len}, references for \texttt{imp}); $\sigma$: variation between draws of the same goal; $n = \sigma^2/\rho^2$: the prediction's weight in draws; $\delta$: the factor a redraw must promise.}
  \label{tab:value-functions}
\end{table}

\textbf{Bins and prompt.} For \texttt{comp}, the seven bins are: under \$0.016, factor-of-two bins from \$0.016 to \$0.512, and over \$0.512; a goal is priced at its bin's geometric center, and the open top bin at \$0.72. For \texttt{len}, they are: under $512$ tokens, factor-of-two bins up to $16{,}384$, and over $16{,}384$. The value function may also answer that the goal is false, which prices it at $\infty$. We use the same model as the prover, DeepSeek-V4-Flash, at low reasoning effort, prompted with the goal's Lean statement and $40$ worked examples: goals from runs on the calibration problems, each labelled with the bin its realized value fell in. \texttt{comp} labels are priced at the $90\%$ cache rate of \S\ref{sec:implementation-setup}, and \texttt{len} labels count tokens as the metric does.

\textbf{Prediction error $\rho$.} We split the calibration problems by problem: the worked examples come from $12$ of them for \texttt{comp} and $11$ for \texttt{len}, and $\rho$ is the root-mean-square of $\log(\text{predicted}/\text{realized})$ over $64$ goals of the remaining $8$ or $7$. For \texttt{comp}, $34\%$ of predictions fall in the right bin and $75\%$ within one bin, and predictions are low by a factor of $1.53$ on average. For \texttt{len}, $40\%$ fall in the right bin and $89\%$ within one, with no systematic bias. For \texttt{imp}, $\rho$ is the root-mean-square impurity of the $125$ solved goals of the $18$ solved calibration problems, \ie, the error of predicting $0$.

\textbf{Spread between draws $\sigma$.} Measuring $\sigma$ requires drawing the same goal repeatedly: on $162$ goals from a held-out set, each redrawn $6$ times, gives $\sigma = 0.615 \pm 0.039$ in log dollars and $0.408$ in log tokens.

\textbf{Redraw margin $\delta$ and scaling.} For \texttt{comp} and \texttt{len}, a redraw must beat the incumbent by one spread, $\delta = e^{\sigma}$. For \texttt{imp}, $e^{\sigma}$ in references is meaningless ($e^{6.79} \approx 890$), so we read the spread off the same draws proportionally: the median ratio of a goal's standard deviation to its mean is $0.72$, giving $\delta = 1.72$. The \texttt{len} estimate is $0.2$ times the bin's center, so the search expects shorter proofs than typical and keeps looking for them; the constant $4$ for \texttt{imp} is the $25$th percentile of impurity over the calibration goals (\S\ref{sec:value-functions}).
\subsection{Cache-Hit Rates and Actual Cost} \label{appdx:realized-cost}

All costs in the main text are re-priced at one common cache-hit rate of 90\% (\S\ref{sec:implementation-setup}). Here we report the rates we observed and the costs they imply.

\begin{table}[h]
    \centering
    \begin{tabular}{l c c}
        \toprule
        Method & Per call & Per token \\
        \midrule
        NearAI    & 93.7\% & 96.2\% \\
        \lever    & 86.2\% & 91.4\% \\
        \bottomrule
    \end{tabular}
    \caption{\textbf{Observed cache-hit rates} on the $80$ evaluation problems. Per call: the mean, over calls, of the fraction of each prompt read from cache. Per token: cached input tokens over all input tokens.}
    \label{tab:cache-rates}
\end{table}

\textbf{Why the rates differ.} Calls went through OpenRouter and InferX, which spread calls over several providers. Further, NearAI carries a single long conversation whereas \lever works on each goal in its own smaller context, reducing cache hit rate.

 \begin{table}[h]
      \centering
      \begin{tabular}{l cc cc}
          \toprule
          & \multicolumn{2}{c}{NearAI} & \multicolumn{2}{c}{\lever} \\
          \cmidrule(lr){2-3} \cmidrule(lr){4-5}
          Pricing & Solved & Cost~(\$) & Solved & Cost~(\$) \\
          \midrule
          90\%, common (Table~\ref{tab:cost-results}) & 0.800 & 1.44 & 0.963 & 0.95 \\
          93\%, common & 0.825 & 1.28 & 0.963 & 0.85 \\
          95\%, common & 0.875 & 1.15 & 0.963 & 0.78 \\
          97\%, common & 0.888 & 1.00 & 0.963 & 0.71 \\
          \midrule
          Own rate per call & 0.850 & 1.24 & 0.950 & 1.09 \\
          Actual & 0.863 & 1.07 & 0.963 & 0.90 \\
          \bottomrule
      \end{tabular}
      \caption{\textbf{Table~\ref{tab:cost-results} under other pricings.} $N=80$, budget \$3. Common: every method at the same rate. Own rate per call: each method at its per-call rate from Table~\ref{tab:cache-rates}.
  Actual: every call at the cache hits it got. Under each pricing, the \$3 rule of Table~\ref{tab:cost-results} is applied to each run's calls in order.}
      \label{tab:cost-pricings}
  \end{table}

\textbf{Costs under other pricings.} Table~\ref{tab:cost-pricings} repeats Table~\ref{tab:cost-results} under each pricing. At any common rate from 90\% to 97\%, \lever solves more problems than NearAI and costs $29$--$34\%$ less. Priced at its own rates, \lever still solves more and costs less: $12\%$ less at its per-call rate, and $16\%$ less at actual cost.

\subsection{Case Studies} \label{appdx:case-study}

\subsubsection{Rejecting a Costly Decomposition: \texttt{putnam\_1972\_a3}} \label{appdx:case-study-1972-a3}

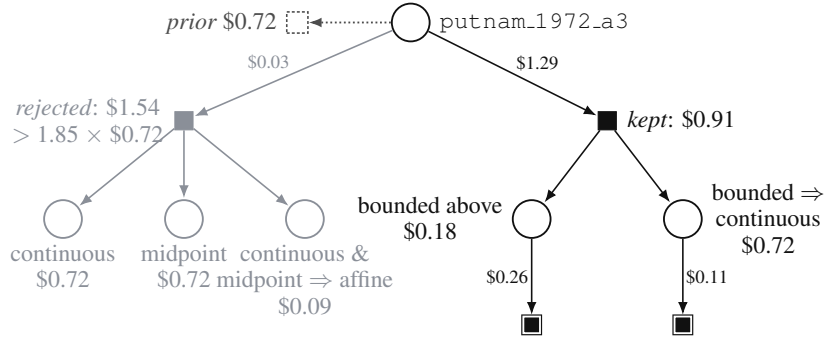
\begin{figure}[h]
  \centering
\begin{tikzpicture}[
    >={Latex[length=1.7mm, width=1.3mm]},
    goal/.style={circle, draw=#1, line width=0.7pt, minimum size=5mm, inner sep=0pt},
    goal/.default=ink,
    att/.style={rectangle, fill=#1, draw=#1, minimum size=2.4mm, inner sep=0pt},
    att/.default=ink,
    solved/.style={att, double=white, double distance=0.7pt, line width=0.45pt, minimum size=2.2mm},
    prior/.style={rectangle, draw=ink!75, dash pattern=on 1.1pt off 0.8pt, line width=0.6pt, minimum size=2.8mm, inner sep=0pt},
    link/.style={->, draw=#1, line width=0.6pt, shorten >=0.4pt},
    link/.default=ink,
    plink/.style={->, draw=ink!75, line width=0.7pt, densely dotted},
    lab/.style={font=\footnotesize, text=ink, inner sep=1pt},
    sub/.style={lab, align=center, text=#1},
    sub/.default=ink,
    cost/.style={font=\scriptsize, text=ink, inner sep=1pt},
    tag/.style={font=\footnotesize, text=#1, inner sep=1pt, align=center},
    tag/.default=muted,
  ]
  \def\ya{0}\def\yb{-1.3}\def\yc{-2.6}\def\yd{-4.0}
  \node[goal] (r) at (0, \ya) {};
  \node[lab, anchor=west] at ([xshift=0.8mm]r.east) {\texttt{putnam\_1972\_a3}};
  \node[prior] (p) at (-1.5, \ya) {};
  \draw[plink] (r) -- (p);
  \node[tag=ink!75, anchor=east] at ([xshift=-0.8mm]p.west) {\textit{prior} \$0.72};
  \node[att=muted] (a1) at (-3.0, \yb) {};
  \draw[link=muted] (r) -- node[cost, text=muted, above left] {\$0.03} (a1);
  \node[tag, anchor=east] at ([xshift=-1mm]a1.west) {\textit{rejected}: \$1.54\\$>$ 1.85 $\times$ \$0.72};
  \node[goal=muted] (g1) at (-4.6, \yc) {};
  \node[goal=muted] (g2) at (-3.0, \yc) {};
  \node[goal=muted] (g3) at (-1.4, \yc) {};
  \foreach \g in {g1, g2, g3} \draw[link=muted] (a1) -- (\g);
  \node[sub=muted, anchor=north] at ([yshift=-0.3mm]g1.south) {continuous\\\$0.72};
  \node[sub=muted, anchor=north] at ([yshift=-0.3mm]g2.south) {midpoint\\\$0.72};
  \node[sub=muted, anchor=north] at ([yshift=-0.3mm]g3.south) {continuous \&\\midpoint $\Rightarrow$ affine\\\$0.09};
  \node[att] (a2) at (2.6, \yb) {};
  \draw[link] (r) -- node[cost, above right] {\$1.29} (a2);
  \node[tag=ink, anchor=west] at ([xshift=1mm]a2.east) {\textit{kept}: \$0.91};
  \node[goal] (g4) at (1.6, \yc) {};
  \node[goal] (g5) at (3.6, \yc) {};
  \draw[link] (a2) -- (g4);
  \draw[link] (a2) -- (g5);
  \node[sub, anchor=east] at ([xshift=-0.8mm]g4.west) {bounded above\\\$0.18};
  \node[sub, anchor=west] at ([xshift=0.8mm]g5.east) {bounded $\Rightarrow$\\continuous\\\$0.72};
  \node[solved] (s4) at (1.6, \yd) {};
  \node[solved] (s5) at (3.6, \yd) {};
  \draw[link] (g4) -- node[cost, left] {\$0.26} (s4);
  \draw[link] (g5) -- node[cost, right] {\$0.11} (s5);
\end{tikzpicture}
  \caption{\textbf{\lever on \texttt{putnam\_1972\_a3}.} Edge labels: what each attempt cost. Subgoal labels: the value function's price. The first decomposition (grey) is rejected before any of its subgoals is attempted.}
  \label{fig:case-1972a3}
\end{figure}

The problem asks for every function on $[0,1]$ that preserves Ces\`aro limits (Figure~\ref{fig:case-1972a3}). The root is priced at \$0.72. The first decomposition leaves three subgoals priced at \$1.54 in total, more than $\delta = 1.85$ times the root's prior, so the search draws the root again without attempting any of them. The second leaves two subgoals priced at \$0.91; it is kept, both are proved, and the run costs \$1.75.
\subsection{Quality Baselines} \label{appdx:quality-baseline}

Both baselines run the agent of \citet{nearaiPutnambenchDeepseek} unchanged, with the model, prices and Lean version of \S\ref{sec:implementation-setup}, and append an objective to the end of its system prompt. So that they are asked for exactly what \lever optimizes, the objective is \lever's own text, word for word: what \lever tells a goal it redraws for that metric. Listing~\ref{lst:objective-impurity} shows the impurity objective, where \texttt{<concepts>} is the statement's subject, read off by the same probe as the metric (Appendix~\ref{appdx:topical-impurity}). The length objective has the same form: it defines a Lean token as the metric counts it, and asks for the shortest complete proof.

\textbf{+ Instruct} solves the problem from its statement, with the objective in its system prompt, and is capped at \$4.5.

\textbf{+ Refactor} starts from NearAI's finished proof of the problem (\S\ref{sec:optimizing-computational-cost}), which replaces the statement in the agent's workspace, and is asked to rewrite it for the objective while keeping it a proof of the same theorem. Its objective adds what \lever tells a redraw about the attempt in hand: for impurity, how many times the proof leaves the subject and which lemmas do, grouped by field; for length, the proof's token count. The run is capped at \$4.5 minus what NearAI's run cost, so that the pipeline as a whole has the same budget as every other method, and its reported cost includes NearAI's.

\begin{lstlisting}[style=prompt, caption={The impurity objective, appended to the baseline agent's system prompt.}, label={lst:objective-impurity}]
# Optimization Objective

`<problem>` is stated in terms of:
```
<concepts>
```
Those are its *subjects*. A proof stays inside the subject when every library fact
it applies is about one of them, or is general enough to belong to general Mathlib
infrastructure; it leaves the subject when it uses declaration(s) belonging to some
other field.

HINTS:
    - Revise the proof to keep it inside the subject the theorem is stated in. A proof
    leaves it when it reaches for a declaration in some other field;
    - If a helper leaves the subject, you should restate the helper so that it stays within the subject;
    - Change any open helper that you believe cannot be closed completely without leaving the subject;
    - You may have to consider a new formalization strategy for the full proof to stay within the subject.
\end{lstlisting}

\section{Theoretical Justification for \lever} \label{appdx:lemma}

Listing~\ref{lst:lemma} proves the exactness claim of \S\ref{sec:adaptive-proof-search-with-lever} in Lean~4, without Mathlib. If $V$ satisfies the backup equations on an acyclic AND/OR graph with nonnegative additive costs, then at every node $n$, $V(n) = k$ exactly when $k$ is the least cost of a solution subgraph rooted at $n$, and $V(n) = \infty$ exactly when none exists (\texttt{lemma1\_exact\_values}). A solution's cost sums edge costs over its tree unfolding, so a shared subgoal counts once per use, and a prior with an exact estimate enters as a childless AND node whose edge costs that estimate. Costs are natural numbers in the formalization.

\lstdefinestyle{leanproof}{
  language=lean4,
  morecomment=[s]{/-}{-/},
  basicstyle=\fontsize{7}{8.4}\fontencoding{T1}\fontfamily{zi4}\selectfont,
  keywordstyle=[1]\bfseries, keywordstyle=[2]\relax,
  commentstyle=\color{muted},
  columns=fullflexible, keepspaces=true, upquote=true, breaklines=true,
  frame=single, framerule=0.4pt, xleftmargin=4pt, xrightmargin=4pt,
  captionpos=b,
  literate={§}{{\S}}1 {¬}{{$\neg$}}1 {·}{{$\cdot$}}1 {→}{{$\rightarrow$}}1
           {↔}{{$\leftrightarrow$}}1 {∀}{{$\forall$}}1 {∃}{{$\exists$}}1
           {∈}{{$\in$}}1 {∞}{{$\infty$}}1 {∧}{{$\wedge$}}1 {≤}{{$\le$}}1
           {▸}{{$\triangleright$}}1 {⟨}{{$\langle$}}1 {⟩}{{$\rangle$}}1,
}
\begin{lstlisting}[style=leanproof, caption={Exact values, verified in Lean~4 (\texttt{Lemma1.lean}).}, label={lst:lemma}]
/-!
# LEVER, Lemma 1 (Exact values)

"For any nonnegative cost that adds along edges, the backed-up value `V(r)` is
exactly the cost of the cheapest solution subgraph of the graph built so far."

Formalization (core Lean 4, no Mathlib):
* Costs are natural numbers (nonnegative by construction); `∞` is `none : Option Nat`.
* The graph is a finite-branching AND/OR DAG; acyclicity is witnessed by a `rank`
  that strictly decreases along every edge.
* A solution subgraph is defined inductively; its cost adds edge costs along every
  occurrence (the tree unfolding), which is what the backup of §3.1 computes.
* `V` is *any* function satisfying the backup equations of §3.1. We prove it equals
  the optimal solution cost at every node (so the backup is also unique), and that a
  backed-up `∞` means no solution subgraph exists.
-/

namespace Lever

inductive Kind where
  | or
  | and
  deriving DecidableEq

structure Graph (N : Type) where
  kind    : N → Kind
  ch      : N → List N
  cost    : N → N → Nat
  rank    : N → Nat
  acyclic : ∀ n d, d ∈ ch n → rank d < rank n

/-! ### Extended naturals `Option Nat` (`none = ∞`) -/

def addE (a : Nat) : Option Nat → Option Nat
  | none   => none
  | some b => some (a + b)

/-- Binary sum on `Option Nat`; `∞` absorbs. -/
def add2 : Option Nat → Option Nat → Option Nat
  | some a, some b => some (a + b)
  | _, _           => none

/-- Binary min on `Option Nat`; `∞` is the identity. -/
def min2 : Option Nat → Option Nat → Option Nat
  | none,   r      => r
  | some a, none   => some a
  | some a, some b => some (min a b)

/-- Sum; empty sum is `0`: a childless AND node is solved. -/
def sumE : List (Option Nat) → Option Nat
  | []      => some 0
  | x :: xs => add2 x (sumE xs)

/-- Minimum; empty min is `∞`: a childless OR node is unsolved. -/
def minE : List (Option Nat) → Option Nat
  | []      => none
  | x :: xs => min2 x (minE xs)

def sumN : List Nat → Nat
  | []      => 0
  | x :: xs => x + sumN xs

/-! ### Solution subgraphs and backed-up values -/

/-- `Sol G n k`: a solution subgraph rooted at `n` exists with total cost `k`. -/
inductive Sol {N : Type} (G : Graph N) : N → Nat → Prop where
  | or  {n d : N} {k : Nat} :
      G.kind n = .or → d ∈ G.ch n → Sol G d k → Sol G n (G.cost n d + k)
  | and {n : N} (k : N → Nat) :
      G.kind n = .and → (∀ d ∈ G.ch n, Sol G d (k d)) →
      Sol G n (sumN ((G.ch n).map fun d => G.cost n d + k d))

/-- The backup equations of §3.1. -/
structure IsBackup {N : Type} (G : Graph N) (V : N → Option Nat) : Prop where
  or_eq  : ∀ n, G.kind n = .or →
    V n = minE ((G.ch n).map fun d => addE (G.cost n d) (V d))
  and_eq : ∀ n, G.kind n = .and →
    V n = sumE ((G.ch n).map fun d => addE (G.cost n d) (V d))

/-! ### Auxiliary lemmas -/

theorem minE_mem : ∀ (l : List (Option Nat)) (k : Nat), minE l = some k → some k ∈ l
  | [], k, h => by simp [minE] at h
  | x :: xs, k, h => by
    cases x with
    | none => simp only [minE, min2] at h; exact List.mem_cons_of_mem _ (minE_mem xs k h)
    | some a =>
      cases hr : minE xs with
      | none => simp [minE, min2, hr] at h; simp [h]
      | some b =>
        simp only [minE, min2, hr, Option.some.injEq] at h
        by_cases hab : a ≤ b
        · simp only [Nat.min_def, if_pos hab] at h; simp [h]
        · simp only [Nat.min_def, if_neg hab] at h; subst h
          exact List.mem_cons_of_mem _ (minE_mem xs b hr)

theorem minE_le : ∀ (l : List (Option Nat)) (a : Nat), some a ∈ l →
    ∃ k, minE l = some k ∧ k ≤ a
  | [], a, h => by simp at h
  | x :: xs, a, h => by
    rcases List.mem_cons.mp h with h | h
    · subst h
      cases hr : minE xs with
      | none => exact ⟨a, by simp [minE, min2, hr], Nat.le_refl _⟩
      | some b => exact ⟨min a b, by simp [minE, min2, hr], Nat.min_le_left _ _⟩
    · obtain ⟨k, hk, hka⟩ := minE_le xs a h
      cases x with
      | none => exact ⟨k, by simp [minE, min2, hk], hka⟩
      | some c =>
        exact ⟨min c k, by simp [minE, min2, hk],
          Nat.le_trans (Nat.min_le_right _ _) hka⟩

theorem sumE_some_all : ∀ (l : List (Option Nat)) (k : Nat), sumE l = some k →
    ∀ x ∈ l, ∃ j, x = some j
  | [], _, _, x, hx => by simp at hx
  | y :: ys, k, h, x, hx => by
    cases y with
    | none => simp [sumE, add2] at h
    | some a =>
      cases hr : sumE ys with
      | none => simp [sumE, add2, hr] at h
      | some b =>
        rcases List.mem_cons.mp hx with hx | hx
        · exact ⟨a, hx⟩
        · exact sumE_some_all ys b hr x hx

theorem sumE_map_of_all {N : Type} (c : N → Nat) (V : N → Option Nat) :
    ∀ l : List N, (∀ d ∈ l, ∃ j, V d = some j) →
    sumE (l.map fun d => addE (c d) (V d)) = some (sumN (l.map fun d => c d + (V d).getD 0))
  | [], _ => rfl
  | d :: ds, h => by
    obtain ⟨j, hj⟩ := h d (List.mem_cons_self ..)
    have ih := sumE_map_of_all c V ds (fun e he => h e (List.mem_cons_of_mem _ he))
    rw [List.map_cons, List.map_cons, sumE, sumN, ih, hj]; rfl

theorem sumN_mono {N : Type} (f g : N → Nat) :
    ∀ l : List N, (∀ d ∈ l, f d ≤ g d) → sumN (l.map f) ≤ sumN (l.map g)
  | [], _ => Nat.le_refl _
  | d :: ds, h => by
    simp only [List.map, sumN]
    exact Nat.add_le_add (h d (List.mem_cons_self ..))
      (sumN_mono f g ds (fun e he => h e (List.mem_cons_of_mem _ he)))

/-! ### Main results -/

section
variable {N : Type} {G : Graph N} {V : N → Option Nat}

/-- Soundness: a finite backed-up value is realized by some solution subgraph. -/
theorem backup_sound (hV : IsBackup G V) :
    ∀ (r : Nat) (n : N), G.rank n < r → ∀ k, V n = some k → Sol G n k
  | 0, _, hr, _, _ => absurd hr (Nat.not_lt_zero _)
  | r + 1, n, hr, k, hk => by
    have ih : ∀ d ∈ G.ch n, ∀ j, V d = some j → Sol G d j := fun d hd =>
      backup_sound hV r d (Nat.lt_of_lt_of_le (G.acyclic n d hd) (Nat.le_of_lt_succ hr))
    cases hkind : G.kind n with
    | or =>
      rw [hV.or_eq n hkind] at hk
      obtain ⟨d, hd, hdk⟩ := List.mem_map.mp (minE_mem _ _ hk)
      cases hVd : V d with
      | none => simp [hVd, addE] at hdk
      | some j =>
        simp only [hVd, addE, Option.some.injEq] at hdk
        subst hdk
        exact Sol.or hkind hd (ih d hd j hVd)
    | and =>
      rw [hV.and_eq n hkind] at hk
      have hall : ∀ d ∈ G.ch n, ∃ j, V d = some j := by
        intro d hd
        obtain ⟨x, hx⟩ := sumE_some_all _ k hk _ (List.mem_map_of_mem hd)
        cases hVd : V d with
        | none => simp [hVd, addE] at hx
        | some j => exact ⟨j, rfl⟩
      rw [sumE_map_of_all (G.cost n) V _ hall, Option.some.injEq] at hk
      subst hk
      refine Sol.and (fun d => (V d).getD 0) hkind (fun d hd => ?_)
      obtain ⟨j, hj⟩ := hall d hd
      exact ih d hd _ (by simp [hj])

/-- Optimality: every solution subgraph costs at least the backed-up value
(in particular the backed-up value is finite whenever a solution exists). -/
theorem backup_le (hV : IsBackup G V) {n : N} {k : Nat} (h : Sol G n k) :
    ∃ v, V n = some v ∧ v ≤ k := by
  induction h with
  | @or n d k hkind hd _ ih =>
    obtain ⟨v, hv, hvk⟩ := ih
    obtain ⟨m, hm, hmk⟩ := minE_le ((G.ch n).map fun d => addE (G.cost n d) (V d))
      (G.cost n d + v) (List.mem_map.mpr ⟨d, hd, by simp [hv, addE]⟩)
    exact ⟨m, (hV.or_eq n hkind).trans hm, Nat.le_trans hmk (Nat.add_le_add_left hvk _)⟩
  | @and n k hkind _ ih =>
    have hall : ∀ d ∈ G.ch n, ∃ j, V d = some j := fun d hd =>
      let ⟨v, hv, _⟩ := ih d hd; ⟨v, hv⟩
    refine ⟨_, (hV.and_eq n hkind).trans (sumE_map_of_all (G.cost n) V _ hall), ?_⟩
    apply sumN_mono
    intro d hd
    obtain ⟨v, hv, hvk⟩ := ih d hd
    simpa [hv] using Nat.add_le_add_left hvk (G.cost n d)

/-- **Lemma 1 (Exact values).** If `V` satisfies the backup equations on an acyclic
AND/OR graph with nonnegative additive costs, then at every node `n` (in particular the
root `r`): `V n = some k` iff `k` is the minimum cost of a solution subgraph rooted at `n`,
and `V n = ∞` iff no solution subgraph rooted at `n` exists. -/
theorem lemma1_exact_values (hV : IsBackup G V) (n : N) :
    (∀ k, V n = some k ↔ (Sol G n k ∧ ∀ k', Sol G n k' → k ≤ k')) ∧
    (V n = none ↔ ∀ k, ¬ Sol G n k) := by
  have sound := fun k => backup_sound hV (G.rank n + 1) n (Nat.lt_succ_self _) k
  refine ⟨fun k => ⟨fun hk => ⟨sound k hk, fun k' h' => ?_⟩, fun ⟨hs, hmin⟩ => ?_⟩,
          ⟨fun hn k hs => ?_, fun hno => ?_⟩⟩
  · obtain ⟨v, hv, hvk⟩ := backup_le hV h'
    rw [hk, Option.some.injEq] at hv; exact hv ▸ hvk
  · obtain ⟨v, hv, hvk⟩ := backup_le hV hs
    have := hmin v (sound v hv)
    rw [hv, Nat.le_antisymm hvk this]
  · obtain ⟨v, hv, _⟩ := backup_le hV hs; rw [hn] at hv; cases hv
  · cases hVn : V n with
    | none => rfl
    | some v => exact absurd (sound v hVn) (hno v)

end
end Lever
\end{lstlisting}

\end{document}